\documentclass{article}
\usepackage{ijcai24}

\usepackage{times}
\usepackage{soul}
\usepackage{url}
\usepackage[hidelinks]{hyperref}
\usepackage[utf8]{inputenc}
\usepackage[small]{caption}
\usepackage{graphicx}
\usepackage{amsmath}
\usepackage{amsthm}
\usepackage{booktabs}
\usepackage{algorithm}
\usepackage[noend]{algorithmic}
\usepackage[switch]{lineno}

\usepackage{todonotes}
\usepackage{amssymb}
\usepackage{mathtools}
\usepackage{microtype}
\usepackage{subfigure}

\newcommand{\FT}[1]{\textcolor{black}{#1}}

\newtheorem{example}{Example}

\newtheorem{property}{Property}
\newtheorem{proposition}{Proposition}
\newtheorem{corollary}{Corollary}
\newtheorem{definition}{Definition}
\newtheorem{conjecture}{Conjecture}

\title{Explaining Arguments’ Strength: Unveiling the Role of Attacks and Supports}

\author{
Xiang Yin$^1$
\and
Nico Potyka$^2$\and
Francesca Toni$^1$
\affiliations
$^1$Department of Computing, Imperial College London, UK\\
$^2$School of Computer Science and Informatics, Cardiff University, UK
\emails
\{xy620, ft\}@imperial.ac.uk,
potykan@cardiff.ac.uk
}

\begin{document}

\maketitle

\begin{abstract}
Quantitatively explaining the strength of arguments under gradual semantics 
has recently received increasing attention.
Specifically, several works in the literature provide quantitative explanations 
by computing the attribution scores of arguments. These works disregard the importance of attacks and supports, even though they play an essential role when explaining arguments' strength.
In this paper, we 
propose a novel theory of \emph{Relation Attribution Explanations (RAEs)},
adapting Shapley values from game theory to  
offer fine-grained insights into the 
role of attacks and supports \FT{in quantitative  bipolar argumentation} towards obtaining the arguments' strength
. 
We show that RAEs satisfy several desirable properties
. 
We also propose a probabilistic algorithm to approximate RAEs efficiently
.
Finally, we 
show
the application value of RAEs 
in 
fraud detection and 
large language models case studies.
\end{abstract}


\section{Introduction}
\label{sec_intro}
Explainable Artificial Intelligence (XAI) has received increasing attention in fields such as finance and healthcare, which demand a reliable and legitimate reasoning process.
Argumentation Frameworks (AFs), \FT{e.g. as first studied in}~\cite{Dung_95},
are promising tools in the XAI field~\cite{mittelstadt2019explaining} due to their transparency and interpretability\FT{, as well as their ability to support reasoning about conflicting information}~\cite{vcyras2021argumentative,albini2020deep,potyka2021interpreting,potyka2023explaining,ayoobi2023sparx}.
In Quantitative Bipolar AFs (QBAFs)~\cite{baroni2015automatic}, 
each argument has a \emph{base score}, and its final \emph{strength} is computed by \emph{gradual semantics} based on the strength of its attackers
and supporters\FT{~\cite{baroni2019fine}}.
\FT{QBAFs can be deployed to support several applications.}
For example, \cite{cocarascu2019extracting} build QBAFs to rate movies by aggregating 
movie reviews. The QBAFs have a hierarchical
structure, where the \FT{goodness of} movies is at the top
and influenced by \FT{arguments about} criteria like the quality of
acting \FT{and} directing. These criteria\FT{/arguments}, in turn,
can be affected by subcriteria\FT{/subarguments} like the performance
of particular actors.
\FT{In this application, t}he base scores of arguments are obtained from reviews via a natural language processing pipeline; finally, a gradual semantics is applied 
to determine
the final strength of  movies as their rating scores.

\FT{While the gradual semantics of a QBAF provides an assessment of arguments (e.g., when using QBAFs for aggregating movie reviews, the rating scores of movies),} 
we may also 
be interested in
an intuitive understanding 
of
the \FT{underlying} reasoning process.
This leads to an interesting research question initially raised by \cite{delobelle2019interpretability}:
\textbf{given an argument of interest (\emph{topic argument}) in a QBAF, how to explain the reasoning outcome (i.e., the strength) of this topic argument?}

\begin{figure}[t]
    \centering
    \includegraphics[width=0.6\columnwidth]{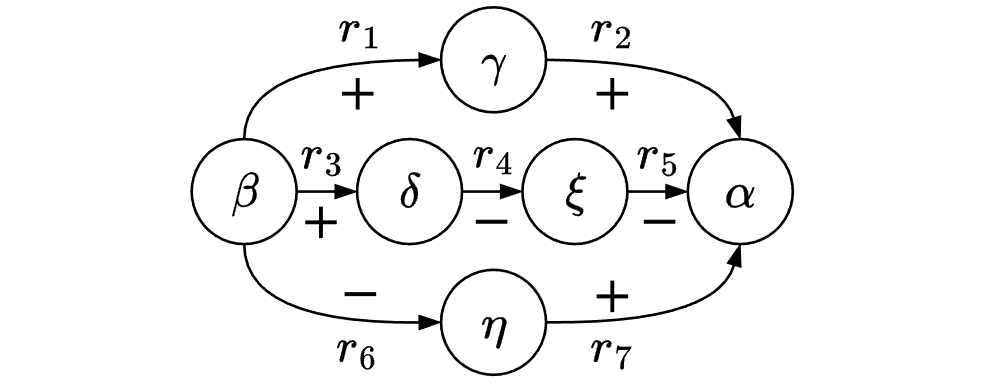}
    \caption{\FT{Graphical view of (elements of) a} QBAF 
   resulting from aggregating movie reviews (here, nodes are arguments, edges labelled + are supports, edges labelled - are attacks, and the $r_i$ are identifiers for the edges (for ease of reference)).
    }
    \label{fig_introduction}
\end{figure}

Most current approaches in the literature 
address this question by defining \emph{argument-based attribution explanations}~\cite{delobelle2019interpretability,vcyras2022dispute,yin2023argument}, which explain the strength of the topic argument by assigning \emph{attribution scores} to arguments: the greater the attribution score, the greater the argument's contribution to the topic argument.
However, 
in many cases,
more fine-grained \emph{relation-based \FT{attribution}
explanations} \FT{(RAEs)}
may be 
beneficial, or even necessary.
For illustration, consider 
Figure~\ref{fig_introduction}, and assume that the QBAF (partially) depicted therein results from aggregating movie reviews
as 
in~\cite{cocarascu2019extracting}, where $\alpha$ is a movie 
to be rated (topic argument)\footnote{We give concrete values for the RAEs in Figure~\ref{fig_introduction} in 
\url{arxiv.org/abs/2404.14304}.
}.
Here, the \emph{review} $\beta$ has a positive argument attribution score by supporting the \emph{famous actor} $\gamma$ and the \emph{influential director} $\delta$, which attacks \emph{bad directing} $\xi$, but this argument view
conceals the fact that \FT{$\beta$}  also weakens $\alpha$
by attacking its \emph{genre} $\eta$, which supports the topic argument.
In contrast, 
(our) RAEs give more fine-grained insights: although $\beta$ has a positive contribution via $r_1$ and $r_3$ to $\alpha$, it also has a negative contribution via $r_6$.

Motivated by the aforementioned 
\FT{considerations, we make the following contributions:}
\begin{itemize}
    \item We propose a novel theory of RAEs (Section~\ref{sec_theory}).
    \item We study desirable properties of RAEs under several gradual semantics (Section~\ref{sec_property}).
    \item We propose a probabilistic algorithm to efficiently approximate RAEs (Section~\ref{sec_approx_algo}).
    \item We carry out two case studies to demonstrate the practical usefulness of RAEs (Section~\ref{sec_case_study}).
\end{itemize}

The proofs of all results are in 
\url{arxiv.org/abs/2404.14304}.

\section{Related Work}
\label{sec_related}


\cite{vcyras2022dispute} propose the general idea of \emph{contribution functions} that compute  quantitative \emph{contributions} from one argument to another under a given
gradual semantics for QBAFs and study three 
such functions, described below.

The \textbf{\emph{removal-based}} contribution function proposed by~\cite{delobelle2019interpretability} measures how the strength of the topic argument changes if an argument is removed. 
In general, removal-based explanations are simple and intuitive for users to understand without a high cognitive burden.
However, a 
 problem with 
 \FT{them} is that removing an argument will also remove paths from its predecessor to the topic argument. 
The measure can therefore overestimate the contribution 
of an argument.
To solve this problem, \cite{delobelle2019interpretability} propose to cut off the direct relations to an argument before removing it, to obtain the mere contribution of this argument.

The \textbf{\emph{gradient-based}} contribution function
captures the \emph{sensitivity} of the topic argument
w.r.t. another argument. 
It is based on the partial derivative of the topic argument's 
strength w.r.t. the base score of another argument.
Arguments with high sensitivity are seen as important.
Following this idea, \cite{yin2023argument} further explored the gradient-based contribution function under the
\emph{DF-QuAD}
gradual semantics~\cite{rago2016discontinuity} and studied its properties in this setting. 

The \textbf{\emph{Shapley-based}} contribution function
uses the Shapley value from coalitional game theory \cite{shapley1951notes} to assign contribution scores.
Each argument in a QBAF is seen as a \emph{player}
that can contribute to the strength of the topic argument.
Although the Shapley-based contribution function is theoretically well-founded, it is significantly harder to compute than
removal and gradient-based methods. 
Our RAEs are based on Shapley values as well, and we work
around the complexity problem by proposing an approximation algorithm.


\FT{Other work focuses on restricted types of QBAFs
. In particular, }
\cite{amgoud2017measuring} propose a contribution function for attack-only Q\FT{B}AFs 
\FT{(where} arguments can only decrease the strength of the topic argument\FT{)} 
\FT{and}  explain an argument's strenght by assigning attribution scores to its \emph{direct} attacks. 
For instance, in Figure~\ref{fig_introduction}, assume \FT{again} that $\alpha$ is the topic argument that needs to be explained.
The \FT{spirit of the} method \FT{is to} attribute the strength of $\alpha$ to $r_{2}$, $r_{5}$ and $r_{7}$, which are directly connected to $\alpha$.\footnote{Here, although Figure~\ref{fig_introduction} is not an attack-only Q\FT{B}AF, we focus on the spirit of the contribution function of \FT{\cite{amgoud2017measuring}}.} 
\FT{Instead,} in our RAEs, we take all edges into account because they all contribute to $\alpha$. Let us take $r_1$ as an example, \FT{representing that $\beta$} 
contributes to the strength of $\alpha$ by strengthening the supporter $\gamma$ of $\alpha$: the contribution of $r_1$ may even be greater than that of
$r_5$ and $r_7$, which means that \emph{indirect} edges 
\FT{may} play an important role
.

\section{Preliminaries}
\label{sec_preliminaries}

To begin with, we recall the definition of QBAFs
. We focus on QBAFs with strength
values in the domain $\mathbb{I}=[0,1]$
\begin{definition}[QBAF]
A \emph{Quantitative Bipolar AF (QBAF)} is a quadruple $\mathcal{Q}=\langle \mathcal{A}, \mathcal{R}^{-}, \mathcal{R}^{+}, \tau \rangle$ where:
\begin{itemize}
\item $\mathcal{A}$ is a set of \emph{arguments};
\item $\mathcal{R}^{-} \subseteq \mathcal{A} \times \mathcal{A}$ is a binary \emph{attack} relation; 
\item $\mathcal{R}^{+} \subseteq \mathcal{A} \times \mathcal{A}$ is a binary \emph{support} relation; 
\item $\mathcal{R}^{-}$ and $\mathcal{R}^{+}$ are disjoint;
\item 
\FT{$\tau\!:\!\mathcal{A} \!\!\rightarrow \!\! \mathbb{I}$ is a function assigning \emph{base scores} to arguments}. 
\end{itemize}
\end{definition}
QBAFs are often denoted graphically (see Figure~\ref{fig_introduction} as an example), where arguments are nodes and edges show the attack or support relations,  labelled by $-$ and $+$, respectively. The base scores can
be seen as  apriori strengths of arguments
when ignoring all other arguments \FT{(and is omitted from graphical representations, as in Figure~\ref{fig_introduction})}. \FT{Seeing QBAFs as graphs allows us to use standard notions such as that of \emph{path}.}

In the remainder, 
unless specified otherwise, we assume as given a generic 
QBAF $\mathcal{Q}\!=\!\langle \mathcal{A}, \mathcal{R}^{-},\! \mathcal{R}^{+}, \!\tau \rangle$ for  $\mathbb{I}\!=\![0,1]$. Also,
we let $\mathcal{R}\!=\!\mathcal{R}^{-} \! \cup \! \mathcal{R}^{+}$ and, for any $\alpha \!\in \!\mathcal{A}$,  
$\mathcal{R}(\alpha)\!=\!\{ (\alpha,\beta) \!\in \!\mathcal{R} \!\mid \! \beta \!\in\! 
\mathcal{A} \}$ denotes the set of all
outgoing edges from 
$\alpha$.

\emph{Gradual semantics} evaluate QBAFs by a function $\sigma: \mathcal{A} \rightarrow \mathbb{I}$ that assigns a (final) \emph{strength} to every argument \FT{(e.g. see} \cite{leite2011social,baroni2015automatic,amgoud2018evaluation}\FT{)}. 
\FT{In most approaches, $\sigma$ is defined via} 
an iterative process that initializes all 
strength values with the base scores and 
then updates the strength values based on
the strength of attackers and supporters. 
The final strength is the limit of this 
process. The 
\FT{process is} guaranteed
to converge for acyclic graphs after at most $n=|\mathcal{A}|$ iterations\footnote{Since the strength of an argument can only be affected by its parents, it actually suffices to update each argument only once by following a topological ordering of the arguments \cite{Potyka19}.}  \cite{Potyka19} and, in practice, 
also quickly converge\FT{s} for cyclic graphs \cite{Potyka18}.
Since we aim to explain the strength, which is only possible when 
it is defined, we will assume convergence for all arguments in the 
remainder, amounting to the following.
\begin{definition}[Well-definedness]
A gradual semantics $\sigma$ is well-defined
for $\mathcal{Q}$ iff $\sigma(\alpha)$ exists for every $\alpha \in \mathcal{A}$.
\end{definition}
\begin{example}
\label{example_1}
Consider the QBAF in Figure~\ref{fig_counter} where the base scores of all arguments are set to $0.5$. 
Then, the DF-QuAD gradual semantics~\cite{rago2016discontinuity}, denoted by $\sigma^{DF}$, determines the following strengths:\footnote{We omit  details on how  gradual semantics determine strengths, as  we focus on explaining these strengths. For details see \url{arxiv.org/abs/2404.14304}.
}
%
$\sigma^{DF}(\alpha)\!\!=\!\!0.8046875$, 
$\sigma^{DF}(\beta)\!\!=\!\!
\sigma^{DF}(\gamma)\!\!=\!\!0.375, 
\sigma^{DF}(\delta)\!\!=\!\!0.25$, 
$\sigma^{DF}(\zeta)\!\!=\!\!0.5$.

\begin{figure}[h]
    \centering
    \includegraphics[width=0.55\columnwidth]{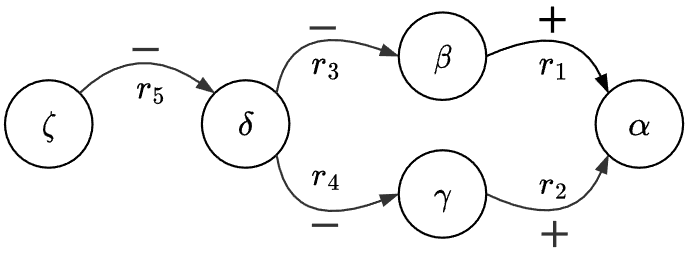}
    \caption{An example of QBAF \FT{(where all base scores are set to $0.5$)}.}
    \label{fig_counter}
\end{figure}
\end{example}




We will often need to restrict
QBAFs to a subset of the edges or change the base score function\FT{, as follows}. 
\begin{definition}
\label{def_sigma_S}
For ${\mathcal{S}} \!\subseteq \! \mathcal{R}$, let $\mathcal{Q}^{|_{\mathcal{S}}} \!\!=\!\!\langle \mathcal{A}, \mathcal{R}^{-}\!\cap \mathcal{S}, \mathcal{R}^{+} \!\cap \mathcal{S}, \tau \rangle$. For
$\tau'\!:\!\mathcal{A} \!\rightarrow \!\mathbb{I}$ a
base score function
, let
$\mathcal{Q}^{|_{\tau'}} \!= \! \langle \mathcal{A}, \mathcal{R}^{-}, \mathcal{R}^{+}, \tau' \rangle$
.
%
Then, 
for any $\alpha\!\in\!  
\mathcal{A}$, we let $\sigma_{\mathcal{S}}(\alpha)$ denote the strength of $\alpha$ in
$\mathcal{Q}^{|_\mathcal{S}}$ and $\sigma_{\tau'}(\alpha)$ denote the strength of $\alpha$ in
$\mathcal{Q}^{|_{\tau'}}$.
\end{definition}
For illustration, in 
Figure~\ref{fig_counter}, suppose $\mathcal{S}=\{r_1,r_3,r_5\}$. Then $r_2$ and $r_4$ are not considered when computing $\sigma_{\mathcal{S}}(\alpha)$.


We will consider the following \emph{monotonicity} property of gradual semantics, which is a variant of various notions proposed in the literature (see \cite{baroni2019fine}
) suitable for our setting. Roughly speaking, it states
that base scores and 
\FT{relations} monotonically influence 
arguments as one would intuitively expect.
\begin{definition}[Monotonicity]
\label{def_mono}
A gradual semantics $\sigma$ is monotonic iff for any $\alpha,\beta \in \mathcal{A}$ such that $\alpha \neq \beta$ and $\mathcal{R}(\beta)=\{(\beta,\alpha)\}$, for any $\tau':\mathcal{A} \rightarrow \mathbb{I}$:
    \begin{enumerate}
        \item 
        
        If $(\beta, \alpha) \in \mathcal{R}^{-}$, then $\sigma(\alpha) \leq
         \sigma_{\mathcal{R} \setminus \{ (\beta, \alpha) \}}(\alpha)$;
        \item 
        
        If $(\beta, \alpha) \in \mathcal{R}^{+}$, then $\sigma(\alpha) \geq
         \sigma_{\mathcal{R} \setminus \{ (\beta, \alpha) \}}(\alpha)$;  
        \item 
        
        If $(\beta, \alpha) \in \mathcal{R}^{-}$, $\tau(\beta) \leq \tau'(\beta)$ and $\tau(\gamma) = \tau'(\gamma)$ for all $\gamma \in \mathcal{A} \setminus \{\beta\}$, then $\sigma(\alpha) \geq \sigma_{\tau'}(\alpha)$;
        \item 
        
        If $(\beta, \alpha) \in \mathcal{R}^{+}$, $\tau(\beta) \leq \tau'(\beta)$ and $\tau(\gamma) = \tau'(\gamma)$ for all $\gamma \in \mathcal{A} \setminus \{\beta\}$, then $\sigma(\alpha) \leq \sigma_{\tau'}(\alpha)$.
    \end{enumerate}
\end{definition}

DF-QuAD, Quadratic Energy (QE)~\cite{Potyka18}, Restricted Euler-based (REB)~\cite{amgoud2018evaluation} and all commonly considered gradual semantics in the literature satisfy monotonicity in acyclic QBAFs. We conjecture that these semantics are also monotonic for cyclic QBAFs.
\begin{proposition}
\label{monotonicity_satisfaction}
DF-QuAD, QE, REB satisfy monotonicity in acyclic QBAFs.
\end{proposition}
\begin{conjecture}
DF-QuAD, QE, REB satisfy monotonicity in cyclic QBAFs.
\end{conjecture}

\section{Relation Attribution Explanations}
\label{sec_theory}
In order to explain the strength of a topic argument in a QBAF, we 
quantify the contribution of
all edges to the topic argument. 
In order to find a fair and reasonable attribution method \FT{for quantifying these contributions}, we 
build up on the Shapley-value as in \cite{amgoud2017measuring,vcyras2022dispute}. We define our \emph{relation attribution explanations} as follows.

\begin{definition}[
RAEs]
\label{def_RAE}
Let $\alpha \in \mathcal{A}$ be a topic argument and $r \in \mathcal{R}$. We define the \emph{Relation Attribution Explanation (RAE) from $r$ to $\alpha$ under $\sigma$} as:
$$
\phi_{\sigma}^{\alpha}(r)\!=\!\!\!\!\!\sum_{\mathcal{S} \subseteq \mathcal{R} \setminus \{r\}}\!\! \frac{(\left| \mathcal{R} \right| - \left| \mathcal{S} \right| -1)!\left| \mathcal{S} \right|!}{\left| \mathcal{R} \right|!} \left[ \sigma_{\mathcal{S} \cup \{r\}}(\alpha)-\sigma_{\mathcal{S}}(\alpha)\right].
$$
\end{definition}
Intuitively, $\phi_{\sigma}^{\alpha}(r)$ looks at every subset of
edges ($\mathcal{S}
$) and computes the marginal
contribution of $r$ ($\left[ \sigma_{\mathcal{S} \cup \{r\}}(\alpha)-\sigma_{\mathcal{S}}(\alpha)\right]$). This marginal contribution is weighted by the probability that a random permutation of the edges starts with the subset ($\mathcal{S}$) and is followed by $r$.
The main difference between our definition and 
\FT{that} in \cite{amgoud2017measuring} lies in the potential ``causes'' of topic arguments. We attribute the strength of a topic argument to all edges (direct and indirect causes) in the QBAF while \cite{amgoud2017measuring} attribute \FT{it} only to the directly incoming edges (direct causes). 
Furthermore, our definition is suitable not only for attacks but also for supports~\cite{cayrol2013bipolarity}, which are important in 
applications~\cite{delobelle2019interpretability}.

Qualitatively, we distinguish three different 
\FT{\emph{(relation) contributions}} based on the sign of $\phi_{\sigma}^{\alpha}(r)$.
\begin{definition}[(Relation) 
\FT{Contribution}]
\label{def_influence}
Let $\alpha \in \mathcal{A}$ and $r \in \mathcal{R}$.
        
        1. If $\phi_{\sigma}^{\alpha}(r) > 0$,  we say $r$ has a \emph{positive contribution to} $\alpha$;
        
        2. If $\phi_{\sigma}^{\alpha}(r) < 0$,  we say $r$ has a \emph{negative contribution to} $\alpha$;
        
        3. If $\phi_{\sigma}^{\alpha}(r) = 0$,  we say $r$ has a \emph{neutral contribution to} $\alpha$.
\end{definition}

\begin{example}[Cont]
\label{example_contribution}
Consider 
again the QBAF in 
\FT{Figure~\ref{fig_counter}} 
\FT{under the} DF-QuAD gradual semantics \FT{as in Example~\ref{example_1}}.
Let $\alpha$ be the topic argument.
We compute RAEs by Definition~\ref{def_RAE}:
$\phi_{\sigma^{DF}}^{\alpha}(r_1) = 0.16875>0$,
$\phi_{\sigma^{DF}}^{\alpha}(r_2)\! = 0.16875>0$,
$\phi_{\sigma^{DF}}^{\alpha}(r_3) \approx -0.0318<0$,
$\phi_{\sigma^{DF}}^{\alpha}(r_4) \approx -0.0318<0$,
$\phi_{\sigma^{DF}}^{\alpha}(r_5) \approx 0.0307>0$.
Hence, $r_1$, $r_2$ and $r_5$ have a positive contribution to $\alpha$ while $r_3$ and $r_4$ have a negative one.
We can also see that $r_1$ and $r_2$ have a more positive contribution to $\alpha$ than $r_5$.
We visualize the RAEs in Figure~\ref{fig_color}. 
\begin{figure}[h]
    \centering
    \includegraphics[width=0.55\columnwidth]{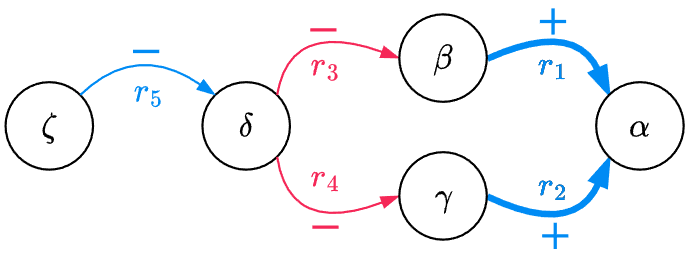}
    \caption{ \FT{Contributions, drawn from} RAEs, for 
    \FT{topic argument $\alpha$ for the QBAF in 
    Figure~\ref{fig_counter}.}
    (Blue/red edges denote \emph{positive}/\emph{negative} contributions, respectively. The thickness of edges represents the magnitude of their contributions\FT{, i.e. their RAE value}.)}
    \label{fig_color}
\end{figure}
\end{example}


\section{Properties}
\label{sec_property}

We now study some 
properties of RAEs.
We start with 
\emph{Shapley-based properties}
that basically 
\FT{adapt to our setting} properties of Shapley 
values and then move to 
\emph{argumentative properties} that \FT{we deem} 
interesting in 
our setting.

\subsection{Shapley-based Properties}

Similar to \cite{amgoud2017measuring}, we first transfer 
the four basic properties of Shapley-values \cite{shapley1951notes} to our setting. 
\emph{Efficiency} is recognized as a desirable property for attribution methods~\cite{ancona2017towards}.
In our context, it states that the sum
of all RAEs corresponds to
the deviation of the topic argument's strength $\sigma(\alpha)$
from its base score $\tau(\alpha)$ (\FT{namely} the explanation distributes 
the responsibility for the difference among the edges). To prove this property, we assume
that the semantics satisfies the \emph{Stability}
property \cite{amgoud2018evaluation}, which states
that the final strength of an argument is its base
score whenever it has no 
\FT{incoming edges}.
\begin{proposition}[Efficiency]
\label{proposition_efficiency}
If $\sigma$ satisfies  \emph{Stability},
then for all $\alpha \in \mathcal{A}$ and $r \in \mathcal{R}$:
$
\sigma(\alpha)=\tau(\alpha)+\sum_{r \in \mathcal{R}}\phi_{\sigma}^{\alpha}(r).
$
\end{proposition}
As an illustration, in Example~\ref{example_contribution}, $\sigma^{DF}(\alpha)=0.8046875$ equals to the the sum of $\tau(\alpha)=0.5$ and the RAEs for all 
five edges \FT{in $\mathcal{Q}$:} $\sum_{r \in \mathcal{R}}\phi_{\sigma^{DF}}^{\alpha}(r) =0.3046875$.

Efficiency has an interesting implication that 
is 
called \emph{Justification} in \cite{vcyras2022dispute}.
\FT{This} demands that whenever the strength of the topic argument differs from its base score, then there is a non-zero RAE 
explaining the difference.
\begin{corollary}[Justification]
\label{proposition_just}
Let $\alpha \in \mathcal{A}$ and $r \in \mathcal{R}$.
    
    1. If $\sigma(\alpha) > \tau(\alpha)$, then $\exists r \in \mathcal{R}$ such that $\phi_{\sigma}^{\alpha}(r) > 0$;
    
    2. If $\sigma(\alpha) < \tau(\alpha)$, then $\exists r \in \mathcal{R}$ such that $\phi_{\sigma}^{\alpha}(r) < 0$.
\end{corollary}
As an illustration, in Example~\ref{example_contribution}, we have $\sigma^{DF}(\alpha)>\tau(\alpha)$ and $r_1$ with $\phi_{\sigma^{DF}}^{\alpha}(r_1) = 0.16875 > 0$ justifies the
difference.

\emph{Dummy}, also known as \emph{Missingness} in \cite{lundberg2017unified}, guarantees that if an edge does not 
\FT{make} any contribution to 
the topic argument, then its RAE is $0$. 
\begin{proposition}[Dummy]
\label{proposition_dummy}
Let $\alpha \in \mathcal{A}$ and $r \in \mathcal{R}$. If $\sigma_{\mathcal{S} \cup \{r\}}(\alpha)=\sigma_{\mathcal{S}}(\alpha)$ holds for all $\mathcal{S} \subseteq \mathcal{R}$, then
$
\phi_{\sigma}^{\alpha}(r)=0.
$
\end{proposition}
As an illustration, in Example~\ref{example_contribution}, if we explain $\sigma^{DF}(\beta)$ \FT{(i.e. the topic argument is $\beta$)}, then $\phi_{\sigma^{DF}}^{\beta}(r_1)=0$.

\emph{Symmetry} states that if two edges share the same contribution to the topic argument, then their RAEs are equal.
\begin{proposition}[Symmetry]
\label{proposition_symmetry}
Let $\alpha \in \mathcal{A}$ and $r_{i},r_{j} \in \mathcal{R}$ \FT{with $r_{i}\neq r_{j}$}.
If $\sigma_{\mathcal{S} \cup \{r_{i}\}}(\alpha)=\sigma_{\mathcal{S} \cup \{r_{j}\}}(\alpha)$ holds for any $\mathcal{S} \subseteq \mathcal{R} \setminus \{r_i,r_j\}$, then $\phi_{\sigma}^{\alpha}(r_{i})=\phi_{\sigma}^{\alpha}(r_{j}).$
\end{proposition}
As an illustration, in Example~\ref{example_contribution}, $r_1$ and $r_2$ have symmetrical effects, thus $\phi_{\sigma^{DF}}^{\alpha}(r_1)=\phi_{\sigma^{DF}}^{\alpha}(r_2)=0.1608$.

\emph{Dominance} states that if one edge always 
\FT{makes} a larger 
contribution than another, then this should be reflected in the magnitude of the RAE.
\begin{proposition}[Dominance]
\label{proposition_dominance}
Let $\alpha \in \mathcal{A}$ and $r_{i},r_{j} \in \mathcal{R}$ \FT{with $r_{i}\neq r_{j}$}.
If $\exists \mathcal{S}' \subseteq \mathcal{R} \setminus \{r_i,r_j\}$ such that $\sigma_{\mathcal{S}' \cup \{r_{i}\}}(\alpha)>\sigma_{\mathcal{S}' \cup \{r_{j}\}}(\alpha)$ and $\forall \mathcal{S}'' \subseteq \mathcal{R} \setminus \{r_i,r_j\}$ ($\mathcal{S}' \neq \mathcal{S}''$) such that $\sigma_{\mathcal{S}'' \cup \{r_{i}\}}(\alpha) \geq \sigma_{\mathcal{S}'' \cup \{r_{j}\}}(\alpha)$, then
$
\phi_{\sigma}^{\alpha}(r_{i})>\phi_{\sigma}^{\alpha}(r_{j}).
$
\end{proposition}
As an illustration, in Example~\ref{example_contribution}, let $\tau(\beta)\!=\!1.0$ while all other \FT{base score}s remain $0.5$. For $\mathcal{S}'\!=\!\{r_3,r_4,r_5\}$ and $\forall \mathcal{S}'' \!\subseteq \!\{r_3,r_4,r_5\} (\mathcal{S}' \!\neq \! \mathcal{S}'')$, we have 
$\sigma_{\mathcal{S}' \cup \{r_{1}\}}(\alpha)>\sigma_{\mathcal{S}' \cup \{r_{2}\}}(\alpha)$ and $\sigma_{\mathcal{S}'' \cup \{r_{1}\}}(\alpha) \geq \sigma_{\mathcal{S}'' \cup \{r_{2}\}}(\alpha)$, thus
$\phi_{\sigma^{DF}}^{\alpha}(r_1)=0.3375 > \phi_{\sigma^{DF}}^{\alpha}(r_2)=0.1292$.


\subsection{Argumentative Properties}

We now study some argumentative properties that 
\FT{we deem}
interesting in our setting.
\FT{When assessing these properties,}
we distinguish three edge types in QBAFs based on the \FT{form and} number of paths 
\FT{to} the topic argument. 
\begin{definition}[Edge Types]
\label{def_edge_type}
Let $\alpha, \beta\FT{, \gamma} \in \mathcal{A}$, $\alpha \neq \beta$
. \FT{Then} 
    \begin{enumerate}
        \item 
        %
        
        \FT{$(\beta, \gamma)$ is a \emph{direct} edge w.r.t. $\alpha$ if $(\beta, \gamma) \in \mathcal{R}$ and there is only one path from $\gamma$ to $\alpha$ in $\mathcal{Q}$ (and $\gamma =\alpha$)};
         \item 
        %
        
        \FT{$(\beta, \gamma)$} is an \emph{indirect} edge w.r.t. $\alpha$ if there is only one path from 
        \FT{$\gamma$} to $\alpha$ \FT{in $\mathcal{Q}$} (and 
     \FT{$\gamma \neq \alpha$});
        \item 
        %
        
        \FT{$(\beta, \gamma)$} is a \emph{multifold} edge w.r.t. $\alpha$ if there is more than one path from 
        \FT{$\gamma$} to $\alpha$ \FT{in $\mathcal{Q}$} \FT{(and 
     $\gamma \neq \alpha$)}.
    \end{enumerate}
\end{definition}

\begin{example}[Cont]
\label{example_three_edge_types}
Given the QBAF in 
\FT{Figure~\ref{fig_counter}},
$r_1$ and $r_2$ are direct edges w.r.t. $\alpha$ as they 
\FT{bring direct support} to $\alpha$; 
$r_3$ and $r_4$ are indirect edges w.r.t. $\alpha$ as they are 
\FT{on single paths} to $\alpha$ \FT{(while not bringing support or attack to it)}; 
$r_5$ is a \emph{multifold} edge w.r.t. $\alpha$ because 
\FT{it starts} two \FT{different} paths ($
r_5,r_3,r_1 
$ and $
r_5,r_4,r_2 
$) 
to $\alpha$.
\end{example}

\FT{The first argumentative property is} \emph{Sign Correctness}, demanding that the sign of an edge
reflects its polarity.
\begin{property}[Sign Correctness]
\label{property_sign_correctness}
Let $\alpha \in \mathcal{A}$ and $r \in \mathcal{R}$.
        
        1. If $r \in \mathcal{R^{-}}$, then $\phi_{\sigma}^{\alpha}(r) \leq 0$;
        
        2. If $r \in \mathcal{R^{+}}$, then $\phi_{\sigma}^{\alpha}(r) \geq 0$.
\end{property}
Naturally, sign correctness cannot be satisfied if
the gradual semantics does not behave in the intended way\FT{, but it does so if it satisfies monotonicity, for direct edges}.
Note that we are defining this and later properties wrt. specific arguments and edges, considering then their satisfaction wrt. classes of edges, e.g. direct edges as in the next result.

\begin{proposition}
\label{proposition_sign_correct_direct}
Let $r$ be a direct edge w.r.t. $\alpha$. 
$\phi_{\sigma}^{\alpha}(r)$ satisfies sign correctness 
if $\sigma$ satisfies monotonicity.
\end{proposition}

\FT{For indirect edges, we need to make several case differentiations, as} the meaning of edges can be inverted along
paths (e.g. an attacker of an attacker, actually serves
as a supporter). 

\begin{proposition}
\label{proposition_sign_correct_indirect}
Let $r$ be an indirect edge w.r.t. $\alpha$. Suppose the path sequence from $r$ to $\alpha$ is $
r, r_{1}, \cdots, r_{n} 
(n \geq 1)$. Let $\lambda= \left| \{r_{1}, \cdots, r_{n} \} \cap \mathcal{R}^{-} \right|$. Then the following statements hold if $\sigma$ satisfies monotonicity.
    
        1. If $r \in \mathcal{R^{-}}$ and $\lambda$ is odd, then $\phi_{\sigma}^{\alpha}(r) \geq 0$;
        
        2. If $r \in \mathcal{R^{-}}$ and $\lambda$ is even, then $\phi_{\sigma}^{\alpha}(r) \leq 0$;
        
        3. If $r \in \mathcal{R^{+}}$ and $\lambda$ is odd, then $\phi_{\sigma}^{\alpha}(r) \leq 0$;
        
        4. If $r \in \mathcal{R^{+}}$ and $\lambda$ is even, then $\phi_{\sigma}^{\alpha}(r) \geq 0$.
\end{proposition}
\begin{example}[Cont]
Consider the QBAF in \FT{Figure~\ref{fig_counter} and the RAEs in} Example~\ref{example_contribution}. $r_1 \in \mathcal{R}^{+}$ is a direct edge w.r.t. $\alpha$, hence $\phi_{\sigma^{DF}}^{\alpha}(r_1) \geq 0$;
while $r_3 \in \mathcal{R}^{-}$ is an indirect edge w.r.t. $\alpha$, and $\lambda$ is $0$ (even), hence $\phi_{\sigma^{DF}}^{\alpha}(r_3) \leq 0$.
\end{example}
Essentially, these results show that RAEs correctly explain the behavior of \FT{direct and indirect} edges under monotonicity
.
\FT{For multifold edges, however, monotonicity may not help.} 
\begin{proposition}
\label{proposition_sign_correct_multifold}
Let $r$ be a multifold edge w.r.t. $\alpha$.
$\phi_{\sigma}^{\alpha}(r)$ may violate sign correctness 
even if $\sigma$ satisfies monotonicity.
\end{proposition}

\emph{Counterfactuality} is a natural property which states that the strength of a topic argument will not be increased (decreased) if an edge with positive (negative) 
\FT{contribution} is removed.
\begin{property}[Counterfactuality]
\label{property_counterfactuality}
Let $\alpha \in \mathcal{A}$ and $r \in \mathcal{R}$.

    1. If $\phi_{\sigma}^{\alpha}(r) < 0$, then $\sigma(\alpha) \leq
 \sigma_{\mathcal{R} \setminus \{ r \}}(\alpha)$;
    
    2. If $\phi_{\sigma}^{\alpha}(r) > 0$, then $\sigma(\alpha) \geq \sigma_{\mathcal{R} \setminus \{ r \}}(\alpha)$.
\end{property}

\begin{proposition}
\label{proposition_counterfactuality_direct}
Let $r$ be a direct or indirect edge w.r.t. $\alpha$.
$\phi_{\sigma}^{\alpha}(r)$ satisfies counterfactuality 
if $\sigma$ satisfies monotonicity.
\end{proposition}


\begin{example}[Cont]
\label{example_property_cf}
Consider the QBAF in \FT{Figure~\ref{fig_counter} and the RAEs in} Example~\ref{example_contribution}.
$r_1$ is a direct edge w.r.t. $\alpha$ and $\phi_{\sigma^{DF}}^{\alpha}(r_1) > 0$. If $r_1$ is removed,
then $\sigma^{DF}(\alpha)=0.8046875 > \sigma^{DF}_{\mathcal{R} \setminus \{ r_1 \}}(\alpha)=0.6875$.
$r_3$ is an indirect edge w.r.t. $\alpha$ and $\phi_{\sigma^{DF}}^{\alpha}(r_3) < 0$. If $r_3$ is removed, then $\sigma^{DF}(\alpha)=0.8046875 < \sigma^{DF}_{\mathcal{R} \setminus \{ r_3 \}}(\alpha)=0.84375$.

\end{example}

\begin{proposition}
\label{proposition_counterfactuality_multifold}
Let $r$ be a multifold edge w.r.t. $\alpha$.
$\phi_{\sigma}^{\alpha}(r)$ may violate counterfactuality 
even if $\sigma$ satisfies monotonicity.
\end{proposition}

From a debugging angle, it is worth exploring how the RAE can be adjusted by a user. We find that the RAE of an edge \FT{$(\beta,\gamma)$} is \FT{closely} related to the base score of its \emph{source argument} \FT{$\beta$}\FT{, in the sense of the properties of \emph{Qualitative Invariability} and \emph{Quantitative Variability} defined below}. 

Qualitative Invariability states that an edge with positive RAE 
 will never 
 \FT{make a}
 negative contribution to the topic argument even if the base score of its source argument 
 \FT{changes}.
\begin{property}[Qualitative Invariability]
\label{property_qua_inv}
Let $\alpha,\beta \!\in \!\mathcal{A}$ and $r \!\in \! \mathcal{R(\beta)}$.
Let $\phi_{\delta}$ denote $\phi_{\sigma}^{\alpha}(r)$ when setting $\tau(\beta)$ to some $\delta \in \mathbb{I}$.

    1. If $\phi_{\sigma}^{\alpha}(r) < 0$, then $\forall \delta \in \mathbb{I}$, $\phi_{\delta} \leq 0$;
    
    2. If $\phi_{\sigma}^{\alpha}(r) > 0$, then $\forall \delta \in \mathbb{I}$, $\phi_{\delta} \geq 0$.
\end{property}

\begin{proposition}
\label{proposition_qua_inv_direct}
Let $r$ be a direct or indirect edge w.r.t. $\alpha$.
$\phi_{\sigma}^{\alpha}(r)$ satisfies qualitative invariability 
if $\sigma$ satisfies monotonicity.
\end{proposition}


\begin{example}[Cont]
\label{example_property_qua}
Consider the QBAF in \FT{Figure~\ref{fig_counter} and the RAEs in} 
Example~\ref{example_contribution}.
Since $r_1$ is a direct edge w.r.t. $\alpha$ and $\phi_{\sigma^{DF}}^{\alpha}(r_1) > 0$, then even if $\tau(\beta)$ is changed to some other value $\delta$, the new RAE $\phi_{\delta} \geq 0$ still holds by Proposition~\ref{proposition_qua_inv_direct}.
\end{example}

\begin{proposition}
\label{proposition_qua_inv_multifold}
Let $r$ be a multifold edge w.r.t. $\alpha$.
$\phi_{\sigma}^{\alpha}(r)$ may violate qualitative invariability 
even if $\sigma$ satisfies monotonicity.
\end{proposition}

Quantitative variability states that the RAE of an edge will not be increased (decreased) if the base score of its source argument is decreased (increased).
\begin{property}[Quantitative Variability]
\label{property_variability}
Let $\alpha,\beta \in \mathcal{A}$, $\alpha \neq \beta$, and $\mathcal{R(\beta)}=\{r\}$.
Let $\phi_{\delta}$ denote $\phi_{\sigma}^{\alpha}(r)$ when setting $\tau(\beta)$ to some $\delta \in \mathbb{I}$.
    
    1. If $\delta < \tau(\beta)$, then $\left| \phi_{\delta} \right| \leq \left| \phi_{\sigma}^{\alpha}(r) \right|$;
    
    2. If $\delta > \tau(\beta)$, then $\left| \phi_{\delta} \right| \geq \left| \phi_{\sigma}^{\alpha}(r) \right|$.
\end{property}
\FT{We assess the satisfaction of this property when $r$ is a direct,  indirect or manifold edge w.r.t. the topic argument.}
\begin{proposition}
\label{proposition_quan_var_direct}
Let $r$ be a direct or indirect edge w.r.t.~$\alpha$.
$\phi_{\sigma}^{\alpha}(r)$ satisfies quantitative variability 
if $\sigma$ satisfies monotonicity.
\end{proposition}


\begin{example}[Cont]
\label{example_property_quan}
Consider the QBAF in \FT{Figure~\ref{fig_counter} and the RAEs in} Example~\ref{example_contribution}.
Since $r_1$ is a direct edge w.r.t. $\alpha$ and $\phi_{\sigma^{DF}}^{\alpha}(r_1) > 0$, 
then if $\tau(\beta)$ is increased to some $\delta>\tau(\beta)$, the new RAE $\phi_{\delta}$ will not decrease by Proposition~\ref{proposition_quan_var_direct}.
\end{example}

\begin{proposition}
\label{proposition_quan_var_multifold}
Let $r$ be a multifold edge w.r.t. $\alpha$.
$\phi_{\sigma}^{\alpha}(r)$ may violate quantitative variability 
even if $\sigma$ satisfies monotonicity.
\end{proposition}

Let us note that many properties may be violated for the multifold case even if the underlying gradual semantics is monotonic. This is because monotonicity only guarantees the \emph{direct} effects of attacks and supports on an argument.
Since a single edge can be involved in a large number of paths
to the topic argument, one cannot make a reasonable demand about
its effect without a long list of case differentiations.
We therefore focus on the special case\FT{s} where there is only
a single path from the argument under investigation to the topic argument \FT{(i.e. the cases with direct and indirect edges)}.

\section{Approximating RAEs Probabilistically}
\label{sec_approx_algo}

\FT{Here,} we look at how we can compute RAEs efficiently.

Computing RAEs involves computing the final strength values 
of arguments. The runtime for computing the
\FT{se} values
depends on the graph structure \FT{of the QBAF} and on the gradual semantics.
Let $n= \left| \mathcal{R} \right|$ and $m = |\mathcal{A}|$.
The strength \FT{values} can be computed in linear time $\mathcal{O}(m+n)$ for acyclic QBAFs \cite[Proposition 3.1]{Potyka19}. For some 
pathological examples of cyclic QBAFs, the strength computation may not converge resulting in infinite runtime \cite{mossakowski2018modular,Potyka19}. However, for randomly
generated cyclic QBAFs, the strength values typically converge
in subquadratic time \cite[Figure 7]{Potyka18}. In fact, if the outdegree
of arguments in the QBAF is not too large, the strength values
are guaranteed to converge in linear time \cite[Proposition 3.3]{Potyka19}. To avoid a large number of case differentiations, we just denote the runtime for computing strength values by
$T(m,n)$ in the following.

\FT{Concerning RAEs,} let us first note that we can compute the
\FT{m exactly} in exponential time by inspecting all
subsets of edges (excluding cases where 
\FT{determining} strength values fails to converge).
\begin{proposition}[Computing RAEs exactly]
\label{proposition_Computational_Complexity_normal}
RAEs can be computed in time $\mathcal{O}(n \cdot 2^{n} \cdot T(m,n))$.
\end{proposition}
For larger 
\FT{QBAFs}, we can apply approximation methods.
It is folklore in algorithmic game theory
that Shapley values can be seen as
expected values.
Under this view, the probability of a subset
${\mathcal{S}} \subseteq \mathcal{R} \setminus \{r\}$ is defined by
$P(\mathcal{S}) = \frac{(\left| \mathcal{R} \right| - \left| {\mathcal{S}} \right| -1)!\left| {\mathcal{S}} \right|!}{\left| \mathcal{R} \right|!}$ and the \emph{marginal contribution} of $\mathcal{S}$ is defined by the function
$m(\mathcal{S}) = \sigma_{{\mathcal{S}} \cup \{r\}}(\alpha)-\sigma_{{\mathcal{S}}}(\alpha)$. 
Our RAEs then correspond to the expected
value $E_P[m]$ of the marginal contribution
function under $P$. This interpretation allows
approximating the Shapley values by the
approximation 
Algorithm~\ref{alg:algorithm}\FT{, returning a dictionary $attribution\_dict$ used iteratively to accumulate  pairs assigning estimate values to edges}.
\begin{algorithm}[t]
\caption{An Approximation Algorithm 
\FT{for} RAEs}
\label{alg:algorithm}
\textbf{Input}: A QBAF $\mathcal{Q}=\left\langle\mathcal{A}, \mathcal{R}^{-}, \mathcal{R}^{+}, \tau \right\rangle$; a gradual semantics $\sigma$; a topic argument $\alpha$; the sample size $N$.\\
\textbf{Output}: Approximate RAEs $attribution\_dict$.

\begin{algorithmic}[1] 
\STATE $attribution\_dict \leftarrow \{\}$ \hfill \% empty dictionary
\FOR {$r$ in $\mathcal{R}$} 
\STATE $sum \leftarrow 0$\\
\WHILE {$N > 0$} 
\STATE ${\mathcal{S}} \leftarrow random\_sample(\mathcal{R}\setminus\{r\})$\\
\STATE $sum \leftarrow sum + \left[\sigma_{{\mathcal{S}} \cup \{r\}}(\alpha)-\sigma_{{\mathcal{S}}}(\alpha)\right]$\\
\STATE $N \leftarrow N - 1$
\ENDWHILE
\STATE $attribution\_dict\left[ r \right] \leftarrow sum / N$
\ENDFOR
\STATE \textbf{return} $attribution\_dict$
\end{algorithmic}
\end{algorithm}
By the \emph{Law of Large Numbers} (e.g., Theorem 5.1 in \cite{gut2005probability}), the estimates converge in probability to the true Shapley values. That is, for every $\epsilon > 0$,
the probability that the estimates deviate by
more than $\epsilon$ from the true value
approaches $0$ as the number of samples
approaches infinity. By the \emph{Central Limit 
Theorem} (e.g., Theorem 5.2 in \cite{gut2005probability}), the distribution of the samples
approaches a normal distribution with mean
equal to the true Shapley values.
This means, in particular, that the estimator
is unbiased.
However,
the variance can be quite large and is better evaluated
empirically. We thus
do not give a precise formula
and simply state the following guarantee\FT{s}.
\begin{proposition}
\label{prop_convergence}
The estimates generated by
Algorithm \ref{alg:algorithm}
converge in probability to the true Shapley
values.
\end{proposition}
\begin{proposition}[Approximating RAEs]
\label{proposition_Computational_Complexity_approx}
If the number of samples for each edge is $N$, then approximate RAEs can be generated in time $\mathcal{O}(n \cdot N \cdot T(m,n))$.
\end{proposition}
In particular, when the 
\FT{QBAF} is acyclic or meets the 
conditions 
on the outdegree of arguments 
in \cite[Proposition 3.3]{Potyka19}, we can compute RAEs in time   $\mathcal{O}(n \!\cdot \! N \!\cdot\! (n\!+\!m))$.

\begin{figure}[t]
    \centering
    \includegraphics[width=0.75\columnwidth]{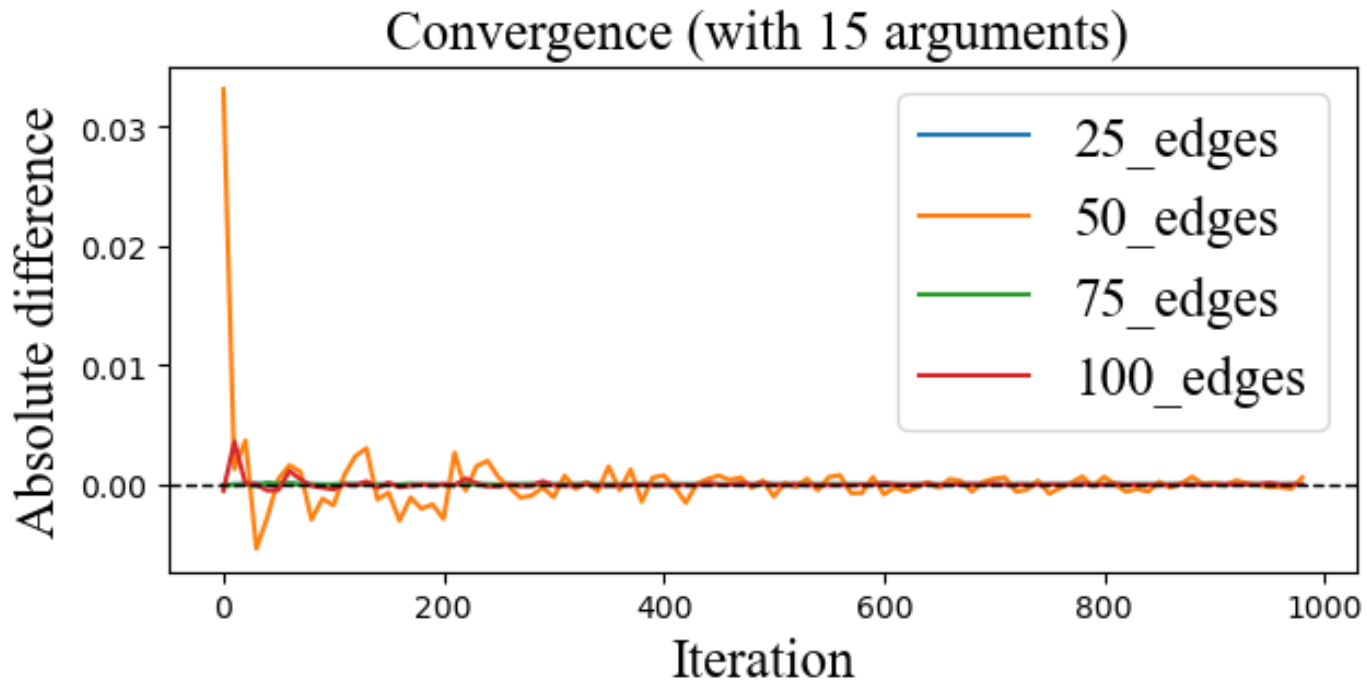}
    \caption{
    Convergence of Algorithm~\ref{alg:algorithm} for random cyclic QBAFs with 15 arguments and 
    various numbers of edges.}
    \label{fig_convergency}
\end{figure}

Proposition \ref{prop_convergence}
guarantees that Algorithm \ref{alg:algorithm} converges
to the true RAEs but does not tell us
how many iterations we require to reach a good approximation.
In order to evaluate the convergence
speed empirically, we conducted
experiments with randomly generated QBAFs of increasing size.
Figure~\ref{fig_convergency}
shows\FT{, for cyclic QBAFs (see 
\url{arxiv.org/abs/2404.14304}
for acyclic QBAFs),} how the absolute difference
(y-axis)
between estimates at every 10-th 
iteration 
evolves with an increasing number
of samples (x-axis)
\FT{, pointing to convergence} within a few hundreds iterations.
For each iteration, it approximately took 14ms and 0.9ms for cyclic and acyclic QBAFs\FT{, respectively,} with 15 arguments and 25 edges.
We give hardware specifications and additional experiments for runtime, acyclic and differently-sized QBAFs in 
\url{arxiv.org/abs/2404.14304}.






\section{Case studies}
\label{sec_case_study}

Finally, we carry out two case studies, including a large QBAF and a non-tree QBAF, to show some practical use of our RAEs
.

\subsection{Case Study 1: Fraud Detection}

\begin{figure}[t]
    \centering
    \includegraphics[width=0.7\columnwidth]{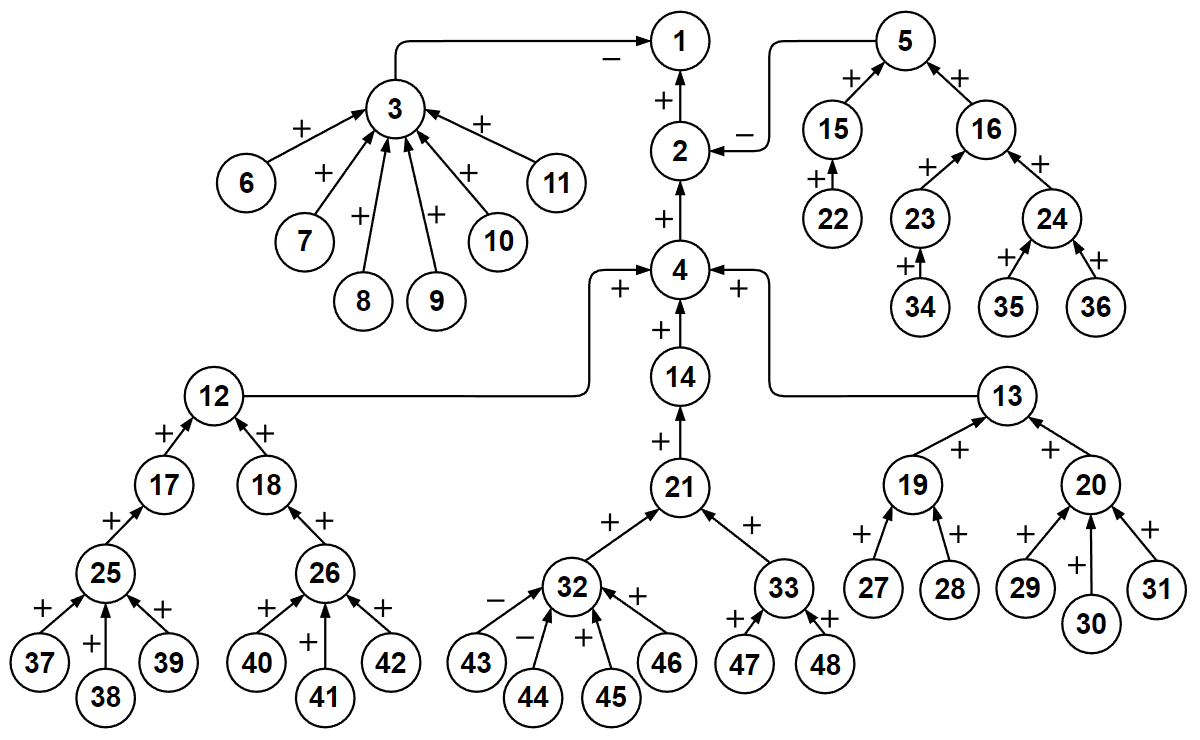}
    \caption{Fraud Detection example from \protect\cite{chi2021optimized}.}
    \label{fig_case1_big_AF}
\end{figure}

\paragraph{Background} 
 Automatic fraud detection plays an important role in e-commerce.
\cite{chi2021optimized} propose to use QBAFs for fraud detection because of their intrinsic interpretability. 
We take the QBAF from \cite{chi2021optimized}, shown in Figure~\ref{fig_case1_big_AF}, where argument $1$ (`It is a fraud case') is the topic argument, and arguments $2-48$ represent evidence for or against this case. The specific content of these arguments can be found in 
\url{arxiv.org/abs/2404.14304}
or in the original
~\cite{chi2021optimized}.

\paragraph{Settings} 
We set the base score for each argument to $0.5$ in line with ~\cite{chi2021optimized}. 
Since we do not consider \FT{in this paper} edge-weighted argumentation~\cite{dunne2011weighted}, we apply DF-QuAD semantics here instead of O-QuAD~\cite{chi2021optimized} which is a variant of DF-QuAD 
\FT{with} weights on edges. 
\FT{The given} case is considered fraud if and only if $\sigma^{DF}(1)>\tau(1)=0.5$. Under DF-QuAD, we have $\sigma^{DF}(1) \approx 0.2544 < \tau(1)$, which means the 
case is not considered fraud
. 
%
Since there are $47$ edges in Figure~\ref{fig_case1_big_AF}, computing RAEs exactly is prohibitively expensive. 
Thus, we apply the approximate Algorithm~\ref{alg:algorithm}, setting the sample size $N$ to $1000$.
We 
chose $N$ experimentally 
to be large enough to guarantee that the estimates converge.\footnote{ 
As the additional experiments for large QBAFs in 
\url{arxiv.org/abs/2404.14304}
show,
the estimates converge typically after a few hundreds iterations even with more than a thousand edges, so $N=1000$ here is appropriate.}

\begin{figure}
    \centering
    \includegraphics[width=0.7\columnwidth]{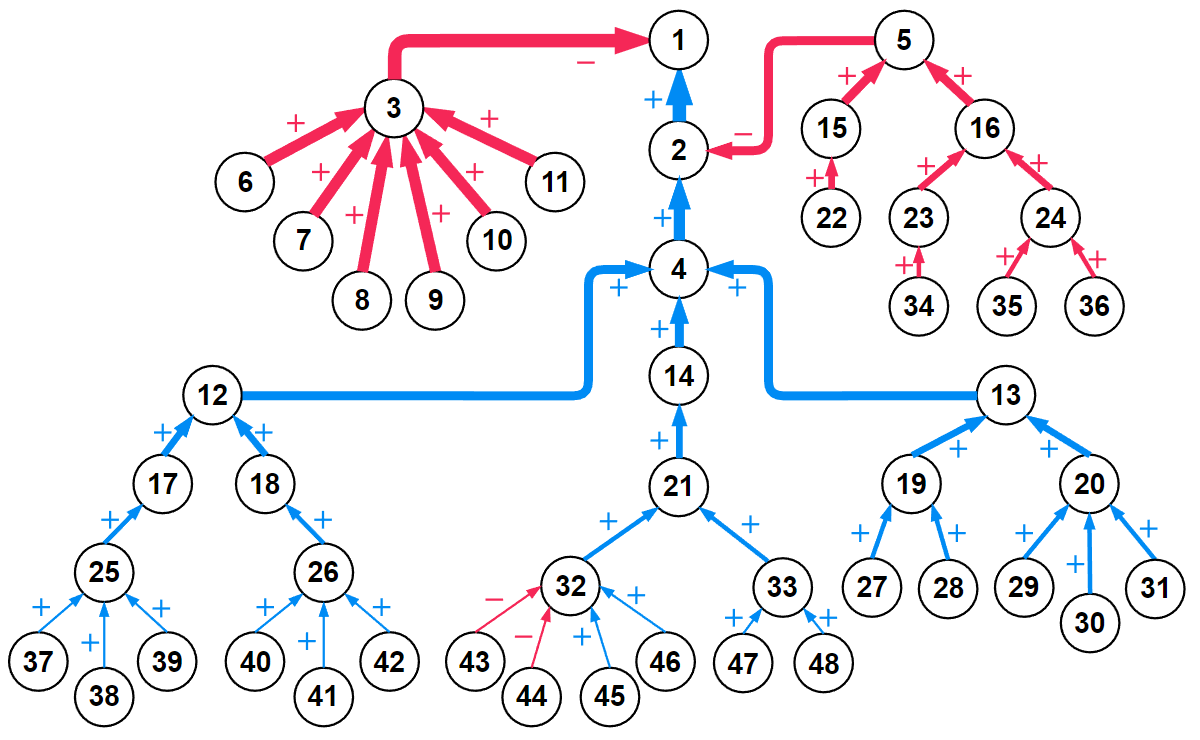}
    \caption{
    \FT{Contributions and} RAEs for the Fraud Detection example. \FT{(We use conventions as described in the caption of Figure~\ref{fig_color})}.}
    \label{fig_case1_big_AF_color}
\end{figure}

\paragraph{Explanations} 
We apply our RAEs \FT{and the contributions derived from them} to 
give quantitative explanations for $\sigma^{DF}(1)$ (see 
Figure~\ref{fig_case1_big_AF_color}; 
for more details see
\url{arxiv.org/abs/2404.14304}
). 

Figure~\ref{fig_case1_big_AF_color} shows that $18$ red edges make a negative contribution while $29$ blue edges make a positive contribution to argument $1$, and negative contributions overwhelm the positive ones.
Among the positive contributions, $(2,1)$ makes the largest, with \FT{RAE=}$2.55 \cdot 10^{-1}$, because it directly supports argument $1$.  $(40,26)$ makes the smallest contribution with \FT{RAE=}$2.83\cdot 10^{-5}$ because it is indirect and far away from argument $1$.
Among the negative contributions, $(3,1)$ makes the largest  (\FT{RAE=}$-4.56\cdot10^{-1}$) since argument $3$ directly attacks argument $1$, whereas $(43,32)$ makes the smallest (\FT{RAE=}$-5.84\cdot10^{-5}$) as there is only one (odd number) attack from argument $43$ to argument $1$.

In Figure~\ref{fig_case1_big_AF_color}, edges close to argument $1$ make a greater contribution than those further away. This is because removing close edges will also remove their predecessors on the
path to the topic argument.
The RAEs of edges such as $(4,2)$ and $(12,4)$ show they also play a role, which is different from the contribution function in \cite{amgoud2017measuring}.

Note that direct edges do not always make greater contribution than those further away, especially in multifold scenarios. 
In such cases, we believe explaining the strength by considering all edges rather than only direct edges 
is a better choice.

\subsection{Case Study 2: Large Language Models (LLMs)}
\paragraph{Background and Settings}
LLMs'
ability to process and generate text can contribute to the development of various AI models~\cite{naveed2023comprehensive} and help address the knowledge acquisition bottleneck.
The idea is that we can query an LLM 
\FT{with a particular claim} and use the answers to build up a QBAF. The QBAF can then be used to visualize (potentially contradictory) arguments that the LLM 
\FT{generated} and compute final strength of these arguments, and RAEs can be used to explain the relevance of particular 
\FT{edges} for the strength of the \FT{claim} (seen as the topic argument).
To present our RAE approach with QBAFs containing more intricate relationships among arguments than the simple tree-like structure resulting in case study 1, we force a certain structure and generate a non-tree QBAF by ChatGPT\FT{(GPT-3.5)}~\cite{chatgpt}, \FT{for the claim 
\emph{`It is easy for children to learn a foreign language well'} (topic argument $\alpha$), prompted to create arguments satisfying the following requirements:}\footnote{
\FT{The prompt 
and response are given in 
\url{arxiv.org/abs/2404.14304.}
}}

\begin{enumerate}
    \item 
    %
    Provide one argument $\beta$ attacking $\alpha$ and two arguments $\gamma$ and $\delta$ supporting $\alpha$.
     \item 
    %
    Let $\beta$ and $\gamma$ attack and support $\delta$, respectively.
     \item 
    %
    \FT{Give} confidence for all arguments, 
    ranging from $0$ to $1$.
\end{enumerate}

\begin{figure}[t]
    \centering
    \includegraphics[width=0.7\columnwidth]{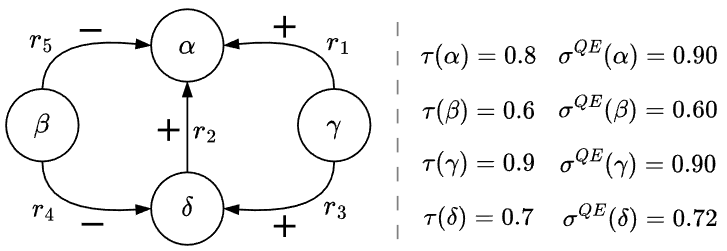}
    \caption{LLMs example.}
    \label{fig_chat1}
\end{figure}

We obtained the following arguments and \FT{confidence values (which we use as} base scores).
\begin{quote}
\emph{\textbf{$\beta$} (0.6): Learning a foreign language requires cognitive maturity, which children lack. Hence, it's difficult for them to excel.}\\
\emph{\textbf{$\gamma$} (0.9): Studies show that young children possess higher neuroplasticity, making language learning more effective.}\\
\emph{\textbf{$\delta$} (0.7): Children immersed in a foreign language environment from an early age have better language acquisition.}    
\end{quote}

We 
used the QE semantics ($\sigma^{QE}$) to compute the strength of arguments
and visualize the  QBAF and 
strengths in Figure~\ref{fig_chat1}.

\begin{figure}[t]
    \centering
    \includegraphics[width=0.65\columnwidth]{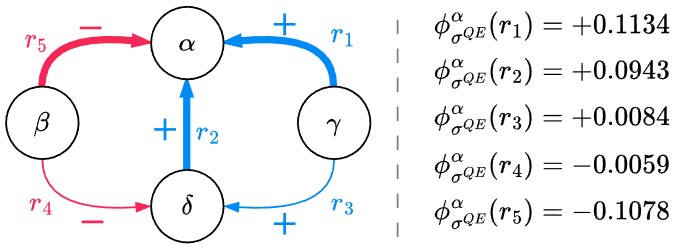}
    \caption{
    \FT{Contributions and} RAEs for the LLMs example 
    .}
    \label{fig_chat2}
\end{figure}

\paragraph{Explanations}
Figure~\ref{fig_chat2} 
\FT{visualises RAEs and contributions} and gives a ranking of the edges based on their contribution.
There are two paths from $\beta$ to $\alpha$: $p_1=r_5$ and $p_2=r_4,r_2$. The \FT{cumulative} contributions of $p_1$ and $p_2$ are $-0.1078$ and $0.0884$, respectively, obtained by adding up the RAEs on the path. Thus, $p_1$ and $p_2$  make, respectively, a negative and positive contribution to $\alpha$. Also, $p_1$ makes a greater contribution 
when considering absolute values.
Although $p_2$ positively contributes to $\alpha$, we find $r_4$ makes a negative contribution on this path, which is not 
\FT{obvious} if we only compute argument-based attributions.
Indeed, we believe that RAEs are better suited than argument-based attribution explanations in this scenario because 
they provide a deeper insight into the way arguments
affect one another along different reasoning paths.


\paragraph{Property Verification}
Let us check \FT{the satisfaction of} some properties introduced previously, under 
$\sigma^{QE}$ \FT{used in this case study}.
First, 
the sum of all RAEs ($0.10$) corresponds to the deviation from $\tau(\alpha)=0.80$ to $\sigma^{QE}(\alpha)=0.90$, which satisfies \emph{efficiency}.
$\beta$ directly attacks $\alpha$ so $r_5$ has a negative RAE while $\gamma$ directly supports $\alpha$ thus $r_1$ has a positive RAE by \emph{sign correctness}. According to \emph{counterfactuality}, if $r_1$ is removed, then $\sigma^{QE}(\alpha)$ will decrease (to $0.80$). If the $\tau(\gamma)$ is increased from $0.9$ to $0.95$, then $\phi_{\sigma^{QE}}^{\alpha}(r_1)$ and $\phi_{\sigma^{QE}}^{\alpha}(r_3)$ are still positive by \emph{qualitative invariability}, and $\phi_{\sigma^{QE}}^{\alpha}(r_1)$ increases from $0.1134$ to $0.1182$ and $\phi_{\sigma^{QE}}^{\alpha}(r_3)$ increases from $0.008389$ to $0.008438$ by \emph{quantitatively variability}.

\section{Conclusion}
\label{sec_conclusion}
We introduced RAEs to quantitatively explain the 
role of 
attack and support relations under gradual semantics for QBAFs, 
\FT{resulting in}  more fine-grained insights into the contribution of arguments, along
different reasoning paths, than argument-based attribution explanations.
We proposed several 
properties for RAEs, including some adapted from \FT{properties of} Shapley values and some defined ex-novo. The satisfaction and violation of these properties theoretically shows that our RAEs are reasonable and faithful explanations, which is crucial to explanation methods. 
We also proposed an efficient probabilistic algorithm to approximate RAEs, proved theoretical convergence guarantees and demonstrated experimentally that it converges quickly. 
Finally, we carried out two 
case studies 
to evaluate and show the practical use of our RAEs.

Our work paves the way to many 
future directions
.
First, it would be interesting to 
explore joint Shapley values~\cite{zhang2021interpreting} for sets of attacks and supports and to investigate interactions among edges.
Second, it would be worth exploring  \FT{formal} relationships between RAEs and argument-based attribution explanations.
Third, it would be interesting to generalize our RAEs to edge-weighted QBAFs
~\cite{amgoud2017measuring}.
Lastly, it would be important to carry out user studies
as explanations should be easily understood and accepted by human users~\cite{chen2022what}.

\section*{Acknowledgments}
This research was partially funded by the  European Research Council (ERC) under the
European Union’s Horizon 2020 research and innovation programme (grant
agreement No. 101020934, ADIX) and by J.P. Morgan and by the Royal
Academy of Engineering under the Research Chairs and Senior Research
Fellowships scheme.  Any views or opinions expressed herein are solely those of the authors.


\bibliography{ijcai24}
\newpage
\setcounter{page}{1}
\onecolumn
\appendix


\section*{Supplementary Material for\\``Explaining Arguments’ Strength: Unveiling the Role of Attacks and Supports''}
\medskip
\section{Computational Details for Gradual Semantics}

\subsection{DF-QuAD Semantics}
\label{dfquad_details}
For any argument $\alpha \in \mathcal{A}$, 
$\sigma^{DF}(\alpha)$ is defined as follows:
$$
\sigma^{DF}(\alpha)= 
    \begin{cases}
        \tau(\alpha)-\tau(\alpha)\cdot(v_{\alpha-att}-v_{\alpha-sup}) & if\ v_{\alpha-att} \geq v_{\alpha-sup}\\
        \tau(\alpha)+(1-\tau(\alpha))\cdot(v_{\alpha-sup}-v_{\alpha-att}) & if\ v_{\alpha-att} < v_{\alpha-sup}\\
    \end{cases}
$$
where $v_{\alpha-att}$ is the aggregation strength of all attackers against $\alpha$, while $v_{\alpha-sup}$ is the aggregation strength of supporters for $\alpha$.
$v_{\alpha-att}$ and $v_{\alpha-sup}$ are defined as follows:
$$
v_{\alpha-att}=1-\prod_{\left \{ \beta \in \mathcal{A} \mid (\beta,\alpha) \in \mathcal{R^{-}} \right \} }(1-\sigma^{DF}(\beta));
$$
$$
v_{\alpha-sup}=1-\prod_{\left \{ \beta \in \mathcal{A} \mid (\beta,\alpha) \in \mathcal{R^{+}} \right \} }(1-\sigma^{DF}(\beta)).
$$
\subsection{QE Semantics}
\label{qe_details}
For any argument $\alpha \in \mathcal{A}$, 
$\sigma^{QE}(\alpha)$ is defined as follows:
$$
\sigma^{QE}(\alpha)= 
    \begin{cases}
        \tau(\alpha)-\tau(\alpha)\cdot\frac{E_\alpha^2}{1+E_\alpha^2} & if\ E_\alpha \leq 0\\
        \tau(\alpha)+(1-\tau(\alpha))\cdot\frac{E_\alpha^2}{1+E_\alpha^2} & if\ E_\alpha > 0\\
    \end{cases}
$$
where $E_\alpha$ is the aggregation strength of all attackers and supporters of $\alpha$. $E_\alpha$ is defined as follows:
$$E_\alpha=\sum_{\left \{ \beta \in \mathcal{A} \mid (\beta,\alpha) \in \mathcal{R^{+}} \right \} }\sigma^{QE}(\beta) - \sum_{\left \{ \beta \in \mathcal{A} \mid (\beta,\alpha) \in \mathcal{R^{-}} \right \} }\sigma^{QE}(\beta).$$

\subsection{REB Semantics}
\label{reb_details}
For any argument $\alpha \in \mathcal{A}$, 
$\sigma^{REB}(\alpha)$ is defined as follows:
$$
\sigma^{REB}(\alpha)=1-\frac{1-\tau^2(\alpha)}{1+\tau(\alpha) \cdot e^{E_\alpha}}
$$
where $E_\alpha$ is the aggregation strength of all attackers and supporters of $\alpha$. $E_\alpha$ is defined as follows:
$$
E_\alpha=\sum_{\left \{ \beta \in \mathcal{A} \mid (\beta,\alpha) \in \mathcal{R^{+}} \right \} }\sigma^{REB}(\beta) - \sum_{\left \{ \beta \in \mathcal{A} \mid (\beta,\alpha) \in \mathcal{R^{-}} \right \} }\sigma^{REB}(\beta).
$$



\section{Proofs for Section~\ref{sec_preliminaries}} 
\setcounter{proposition}{0}
\begin{proposition}
\label{monotonicity_satisfaction}
DF-QuAD, QE, REB satisfy monotonicity in acyclic QBAFs.
\end{proposition}
\begin{proof}
Proof for \textbf{DF-QuAD} semantics:

Let $v_{\mathcal{R} \setminus \{ (\beta, \alpha) \} \alpha-att}$ and $v_{\mathcal{R} \setminus \{ (\beta, \alpha) \} \alpha-sup}$ denote the aggregation value of $\alpha$'s attackers and supporters after removing $(\beta,\alpha)$, respectively.
Let $v'_{\alpha-sup}$ and $v'_{\alpha-att}$ denote the aggregation value of $\alpha$'s attackers and supporters based on the base score function $\tau'$, respectively.

1. Since $\mathcal{R}(\beta)=\{(\beta,\alpha)\}$, if edge $(\beta,\alpha) \in \mathcal{R}^{-}$ is removed, then $\alpha$ loses an attacker $\beta$ and the strength of all other attackers and supporters of $\alpha$ remain. Hence, $v_{\alpha-sup}=v_{\mathcal{R} \setminus \{ (\beta, \alpha) \} \alpha-sup}$ and $v_{\alpha-att} \geq v_{\mathcal{R} \setminus \{ (\beta, \alpha) \} \alpha-att}$ according to the DF-QuAD in Section~\ref{dfquad_details}.

In order to see how $\sigma^{DF}(\alpha)$ changes, we need to discuss the relative magnitude of $v_{\alpha-att}$ and $v_{\alpha-sup}$.
There are three cases:\\
(a) $v_{\alpha-att} \geq v_{\alpha-sup}$ and $v_{\mathcal{R} \setminus \{ (\beta, \alpha) \} \alpha-att} \geq v_{\mathcal{R} \setminus \{ (\beta, \alpha) \} \alpha-sup}$.\\
In this case, $\sigma^{DF}(\alpha)= \tau(\alpha)-\tau(\alpha)\cdot(v_{\alpha-att}-v_{\alpha-sup})$ and $\sigma^{DF}_{\mathcal{R} \setminus \{ (\beta, \alpha) \}}(\alpha) = \tau(\alpha)-\tau(\alpha)\cdot(v_{\mathcal{R} \setminus \{ (\beta, \alpha) \} \alpha-att} - v_{\mathcal{R} \setminus \{ (\beta, \alpha) \} \alpha-sup})$, therefore, we have $\sigma^{DF}(\alpha) \geq \sigma^{DF}_{\mathcal{R} \setminus \{ (\beta, \alpha) \}}(\alpha)$.\\
(b) $v_{\alpha-att} \geq v_{\alpha-sup}$ and $v_{\mathcal{R} \setminus \{ (\beta, \alpha) \} \alpha-att} \leq v_{\mathcal{R} \setminus \{ (\beta, \alpha) \} \alpha-sup}$.\\
In this case, $\sigma^{DF}(\alpha)= \tau(\alpha)-\tau(\alpha)\cdot(v_{\alpha-att}-v_{\alpha-sup})$ and $\sigma^{DF}_{\mathcal{R} \setminus \{ (\beta, \alpha) \}}(\alpha) = \tau(\alpha)+(1-\tau(\alpha))\cdot(v_{\mathcal{R} \setminus \{ (\beta, \alpha) \} \alpha-sup}-v_{\mathcal{R} \setminus \{ (\beta, \alpha) \} \alpha-att})$, therefore, we have $\sigma^{DF}(\alpha) \geq \sigma^{DF}_{\mathcal{R} \setminus \{ (\beta, \alpha) \}}(\alpha)$.\\
(c) $v_{\alpha-att} \leq v_{\alpha-sup}$ and $v_{\mathcal{R} \setminus \{ (\beta, \alpha) \} \alpha-att} \leq v_{\mathcal{R} \setminus \{ (\beta, \alpha) \} \alpha-sup}$.\\
In this case, $\sigma^{DF}(\alpha) = \tau(\alpha)+(1-\tau(\alpha))\cdot(v_{\alpha-sup}-v_{\alpha-att})$ and $\sigma^{DF}_{\mathcal{R} \setminus \{ (\beta, \alpha) \}}(\alpha) = \tau(\alpha)+(1-\tau(\alpha))\cdot(v_{\mathcal{R} \setminus \{ (\beta, \alpha) \} \alpha-sup}-v_{\mathcal{R} \setminus \{ (\beta, \alpha) \} \alpha-att})$, therefore, we have $\sigma^{DF}(\alpha) \geq \sigma^{DF}_{\mathcal{R} \setminus \{ (\beta, \alpha) \}}(\alpha)$.\\
In summary, in any of these three cases, we know that if $(\beta, \alpha) \in \mathcal{R}^{-}$, then $\sigma^{DF}(\alpha) \leq \sigma^{DF}_{\mathcal{R} \setminus \{ (\beta, \alpha) \}}(\alpha)$.

2. Analogously to the proof in 1.

3. Since $\mathcal{R}(\beta)=\{(\beta,\alpha)\}$ and $(\beta, \alpha) \in \mathcal{R}^{-}$, if $\tau(\beta) \leq \tau'(\beta)$ and $\tau(\gamma) = \tau'(\gamma)$ for all $\gamma \in \mathcal{A} \setminus \{\beta\}$,
then $\sigma^{DF}(\beta) \leq \sigma^{DF}_{\tau'}(\beta)$ and $\sigma^{DF}(\gamma) = \sigma^{DF}_{\tau'}(\gamma)$ for all $\gamma \in \mathcal{A} \setminus \{\alpha,\beta\}$. Thus, $v'_{\alpha-att} \geq v_{\alpha-att}$ and $v'_{\alpha-sup} = v_{\alpha-sup}$.

Similarly, in order to see how $\sigma^{DF}(\alpha)$ changes, we need to discuss the relative magnitude of $v_{\alpha-att}$ and $v_{\alpha-sup}$.
There are three cases:\\
(a) $v_{\alpha-att} \geq v_{\alpha-sup}$ and $v'_{\alpha-att} \geq v'_{\alpha-sup}$.\\
In this case, $\sigma^{DF}(\alpha) = \tau(\alpha)-\tau(\alpha)\cdot(v_{\alpha-att}-v_{\alpha-sup})$ and $\sigma^{DF}_{\tau'}(\alpha) = \tau(\alpha)-\tau(\alpha)\cdot(v'_{\alpha-att} - v'_{\alpha-sup})$, therefore, we have $\sigma^{DF}(\alpha) \geq \sigma^{DF}_{\tau'}(\alpha)$.\\
(b) $v_{\alpha-att} \geq v_{\alpha-sup}$ and $v'_{\alpha-att} \leq v'_{\alpha-sup}$.\\
In this case, $\sigma^{DF}(\alpha) = \tau(\alpha)-\tau(\alpha)\cdot(v_{\alpha-att}-v_{\alpha-sup})$ and $\sigma^{DF}_{\tau'}(\alpha) = \tau(\alpha)+(1-\tau(\alpha))\cdot(v'_{\alpha-sup}-v'_{\alpha-att})$, therefore, we have $\sigma^{DF}(\alpha) \geq \sigma^{DF}_{\tau'}(\alpha)$.\\
(c) $v_{\alpha-att} \leq v_{\alpha-sup}$ and $v'_{\alpha-att} \leq v'_{\alpha-sup}$.\\
In this case, $\sigma^{DF} = \tau(\alpha)+(1-\tau(\alpha))\cdot(v_{\alpha-sup}-v_{\alpha-att})$ and $\sigma^{DF}_{\tau'}(\alpha) = \tau(\alpha)+(1-\tau(\alpha))\cdot(v'_{\alpha-sup}-v'_{\alpha-att})$, therefore, we have $\sigma^{DF}(\alpha) \geq \sigma^{DF}_{\tau'}(\alpha)$.\\
In summary, in any of these three cases, we know that if $(\beta, \alpha) \in \mathcal{R}^{-}$, $\tau(\beta) \leq \tau'(\beta)$ and $\tau(\gamma) = \tau'(\gamma)$ for all $\gamma \in \mathcal{A} \setminus \{\beta\}$, then $\sigma^{DF}(\alpha) \leq \sigma^{DF}_{\mathcal{R} \setminus \{ (\beta, \alpha) \}}(\alpha)$.

4. Analogously to the proof in 3.

The proof for \textbf{QE} and \textbf{REB} semantics are analogous to the proof for DF-QuAD semantics.
\end{proof}

\section{Proofs for Section~\ref{sec_property}} 
\subsection{Shapley-based Properties} 

\setcounter{property}{0}
\setcounter{corollary}{0}
\setcounter{example}{0}

\begin{proposition}[Efficiency]
\label{proposition_efficiency}
If $\sigma$ satisfies  \emph{Stability},
then for all $\alpha \in \mathcal{A}$ and $r \in \mathcal{R}$:
$$
\sigma(\alpha)=\tau(\alpha)+\sum_{r \in \mathcal{R}}\phi_{\sigma}^{\alpha}(r).
$$
\end{proposition}

\begin{proof}
According to Definition~\ref{def_RAE}, for any $r \in \mathcal{R}$ we have 
$\phi_{\sigma}^{\alpha}(r)=\sum_{\mathcal{S} \subseteq \mathcal{R} \setminus \{r\}}\!\! \frac{(\left| \mathcal{R} \right| - \left| \mathcal{S} \right| -1)!\left| \mathcal{S} \right|!}{\left| \mathcal{R} \right|!} \left[ \sigma_{\mathcal{S} \cup \{r\}}(\alpha)-\sigma_{\mathcal{S}}(\alpha)\right]$.
This definition is the compact form of the
Shapley value. In order to prove the claim,
we use the equivalent permutation
representation \cite{maschler_solan_zamir_2013}.
\begin{equation}
\phi_{\sigma}^{\alpha}(r)=\frac{1}{\left|\mathcal{R}\right|!}\sum_{\pi \in \Pi(\mathcal{R})}\left[ \sigma_{\pi^{-1}(r) \cup \left\{ r \right\}}(\alpha) - \sigma_{\pi^{-1}(r)}(\alpha)\right],   
\end{equation}
where $\Pi(\mathcal{R})$ denotes the set of all permutations of $\mathcal{R}$ and,
for all $r \in \mathcal{R}$, 
$\pi^{-1}(r)$ denotes the set of edges in $\mathcal{R}$ which precede $r$ in the order $\pi$.
Then, we have
\begin{equation}
\begin{aligned}
\sum_{r \in \mathcal{R}}\phi_{\sigma}^{\alpha}(r)
&=\sum_{r \in \mathcal{R}}\frac{1}{\left|\mathcal{R}\right|!}\sum_{\pi \in \Pi(\mathcal{R})}\left[ \sigma_{\pi^{-1}(r) \cup \left\{ r \right\}}(\alpha) - \sigma_{\pi^{-1}(r)}(\alpha)\right]\\
&=\frac{1}{\left|\mathcal{R}\right|!}\sum_{r \in \mathcal{R}}\sum_{\pi \in \Pi(\mathcal{R})}\left[ \sigma_{\pi^{-1}(r) \cup \left\{ r \right\}}(\alpha) - \sigma_{\pi^{-1}(r)}(\alpha)\right]\\
&=\frac{1}{\left|\mathcal{R}\right|!}\sum_{\pi \in \Pi(\mathcal{R})}\sum_{r \in \mathcal{R}}\left[ \sigma_{\pi^{-1}(r) \cup \left\{ r \right\}}(\alpha) - \sigma_{\pi^{-1}(r)}(\alpha)\right]\\
\end{aligned}
\end{equation}
Most terms in the inner sum cancel.
To see this, consider an arbitrary
permutation $\pi$ and, for ease of notation, assume that it induces
the ordering $(r_1, r_2, \cdots, r_n )$.
Then, we have
\begin{equation}
\begin{aligned}
\label{equ_1}
\sum_{r \in \mathcal{R}}\left[ \sigma_{\pi^{-1}(r) \cup \left\{ r \right\}}(\alpha) - \sigma_{\pi^{-1}(r)}(\alpha)\right]
&=\sigma_{\left\{ r_1 \right\}}(\alpha) - \sigma_{\varnothing}(\alpha)\\
&+\sigma_{ \left\{ r_1,r_2 \right\}}(\alpha) - \sigma_{\left\{ r_1 \right\}}(\alpha)\\
&+\sigma_{ \left\{ r_1,r_2,r_3 \right\}}(\alpha) - \sigma_{\left\{ r_1,r_2 \right\}}(\alpha)\\
&+\cdots\\
&+\sigma_{\left\{ r_1,r_2,r_3, \cdots, r_{n-1} \right\}}(\alpha) - \sigma_{\left\{ r_1,r_2,r_3\cdots,r_{n-2} \right\}}(\alpha)\\
&+\sigma_{\left\{ r_1,r_2,r_3, \cdots, r_n \right\}}(\alpha) - \sigma_{\left\{ r_1,r_2,r_3\cdots,r_{n-1} \right\}}(\alpha)\\
&=\sigma_{\left\{ r_1,r_2,r_3, \cdots, r_n \right\}}(\alpha) - \sigma_{\varnothing}(\alpha)\\
&=\sigma(\alpha) - \tau(\alpha), \\
\end{aligned}
\end{equation}
where the last equality follows from
\emph{Stability}.
Since there are $\left|\mathcal{R}\right|!$ different orderings and $\pi$ is taken arbitrarily, we have

\begin{equation}
\begin{aligned}
\sum_{r \in \mathcal{R}}\phi_{\sigma}^{\alpha}(r)
&=\frac{1}{\left|\mathcal{R}\right|!}\sum_{\pi \in \Pi(\mathcal{R})}\left[ \sigma(\alpha) - \tau(\alpha)\right]\\
&=\frac{1}{\left|\mathcal{R}\right|!}\left|\mathcal{R}\right|!\left[ \sigma(\alpha) - \tau(\alpha)\right]\\
&=\sigma(\alpha) - \tau(\alpha)
\end{aligned}
\end{equation}

Thus, $\sigma(\alpha)=\tau(\alpha)+\sum_{r \in \mathcal{R}}\phi_{\sigma}^{\alpha}(r)$, which completes the proof.

\end{proof}

\begin{corollary}[Justification]
\label{proposition_just}
Let $\alpha \in \mathcal{A}$ and $r \in \mathcal{R}$.
\begin{enumerate}
    \item If $\sigma(\alpha) > \tau(\alpha)$, then $\exists r \in \mathcal{R}$ such that $\phi_{\sigma}^{\alpha}(r) > 0$;
    \item If $\sigma(\alpha) < \tau(\alpha)$, then $\exists r \in \mathcal{R}$ such that $\phi_{\sigma}^{\alpha}(r) < 0$.
\end{enumerate}

\end{corollary}
\begin{proof}
1. According to Proposition~\ref{proposition_efficiency}, $\sigma(\alpha) > \tau(\alpha)$ means $\sum_{r \in \mathcal{R}}\phi_{\sigma}^{\alpha}(r)>0$, hence there must be at least one $r \in \mathcal{R}$ such that $\phi_{\sigma}^{\alpha}(r) > 0$. \\
2. According to Proposition~\ref{proposition_efficiency}, $\sigma(\alpha) < \tau(\alpha)$ means $\sum_{r \in \mathcal{R}}\phi_{\sigma}^{\alpha}(r)<0$, hence there must be at least one $r \in \mathcal{R}$ such that $\phi_{\sigma}^{\alpha}(r) < 0$. 
\end{proof}

\begin{proposition}[Dummy]
\label{proposition_dummy}
Let $\alpha \in \mathcal{A}$ and $r \in \mathcal{R}$. If $\sigma_{\mathcal{S} \cup \{r\}}(\alpha)=\sigma_{\mathcal{S}}(\alpha)$ holds for any $\mathcal{S} \subseteq \mathcal{R} \setminus \{r\}$, then
$$
\phi_{\sigma}^{\alpha}(r)=0.
$$
\end{proposition}
\begin{proof}
According to Definition~\ref{def_RAE}, if  $\sigma_{\mathcal{S} \cup \{r\}}(\alpha)=\sigma_{\mathcal{S}}(\alpha)$ holds for any $\mathcal{S} \subseteq \mathcal{R} \setminus \{r\}$, then $\sigma_{\mathcal{S} \cup \{r\}}(\alpha)-\sigma_{\mathcal{S}}(\alpha)=0$. Thus, $\sum_{\mathcal{S} \subseteq \mathcal{R} \setminus \{r\}} \frac{(\left| \mathcal{R} \right| - \left| \mathcal{S} \right| -1)!\left| \mathcal{S} \right|!}{\left| \mathcal{R} \right|!} \left[ \sigma_{\mathcal{S} \cup \{r\}}(\alpha)-\sigma_{\mathcal{S}}(\alpha)\right]=0$, hence $\phi_{\sigma}^{\alpha}(r)=0$, which completes the proof.
\end{proof}

\begin{proposition}[Symmetry]
\label{proposition_symmetry}
Let $\alpha \in \mathcal{A}$ and $r_{i},r_{j} \in \mathcal{R}$ with $r_i \neq r_j$.
If $\sigma_{\mathcal{S} \cup \{r_{i}\}}(\alpha)=\sigma_{\mathcal{S} \cup \{r_{j}\}}(\alpha)$ holds for any $\mathcal{S} \subseteq \mathcal{R} \setminus \{r_i,r_j\}$, then $$\phi_{\sigma}^{\alpha}(r_{i})=\phi_{\sigma}^{\alpha}(r_{j}).$$
\end{proposition}
\begin{proof}
According to Definition~\ref{def_RAE}, if for any $\mathcal{S} \subseteq \mathcal{R} \setminus \{r_i,r_j\}$, 
we have $\sigma_{\mathcal{S} \cup \{r_{i}\}}(\alpha)=\sigma_{\mathcal{S} \cup \{r_{j}\}}(\alpha)$, then we have $\sigma_{\mathcal{S} \cup \{r_{i},r_{j}\}}(\alpha)-\sigma_{\mathcal{S} \cup \{r_{i}\}}(\alpha)=\sigma_{\mathcal{S} \cup \{r_{j},r_{i}\}}(\alpha)-\sigma_{\mathcal{S} \cup \{r_{j}\}}(\alpha)$. 
We let $\mathcal{S}_{i}=\mathcal{S} \cup \{ r_{i} \}$ and $\mathcal{S}_{j}=\mathcal{S} \cup \{ r_{j} \}$. 
Thus, $\sigma_{\mathcal{S}_{i} \cup \{r_{j}\}}(\alpha)-\sigma_{\mathcal{S}_{i}}(\alpha)=\sigma_{\mathcal{S}_{j} \cup \{r_{i}\}}(\alpha)-\sigma_{\mathcal{S}_{j}}(\alpha)$. 
Therefore, $\sum_{\mathcal{S}_{i} \subseteq \mathcal{R} \setminus \{r_{j}\}} \frac{(\left| \mathcal{R} \right| - \left| \mathcal{S}_{i}\right| -1)!\left| \mathcal{S}_{i} \right|!}{\left| \mathcal{R} \right|!} \left[ \sigma_{\mathcal{S}_{i} \cup \{r_{j}\}}(\alpha)-\sigma_{\mathcal{S}_{i}}(\alpha) \right]=\sum_{\mathcal{S}_{j} \subseteq \mathcal{R} \setminus \{r_{i}\}} \frac{(\left| \mathcal{R} \right| - \left| \mathcal{S}_{j}\right| -1)!\left| \mathcal{S}_{j} \right|!}{\left| \mathcal{R} \right|!} \left[ \sigma_{\mathcal{S}_{j} \cup \{r_{i}\}}(\alpha)-\sigma_{\mathcal{S}_{j}}(\alpha) \right]$, hence $\phi_{\sigma}^{\alpha}(r_{i})=\phi_{\sigma}^{\alpha}(r_{j})$, which completes the proof.
\end{proof}

\begin{proposition}[Dominance]
\label{proposition_dominance}
Let $\alpha \in \mathcal{A}$ and $r_{i},r_{j} \in \mathcal{R}$ with $r_i \neq r_j$.
If $\exists \mathcal{S}' \subseteq \mathcal{R} \setminus \{r_i,r_j\}$ such that $\sigma_{\mathcal{S}' \cup \{r_{i}\}}(\alpha)>\sigma_{\mathcal{S}' \cup \{r_{j}\}}(\alpha)$ and $\forall \mathcal{S}'' \subseteq \mathcal{R} \setminus \{r_i,r_j\}$ ($\mathcal{S}' \neq \mathcal{S}''$) such that $\sigma_{\mathcal{S}'' \cup \{r_{i}\}}(\alpha) \geq \sigma_{\mathcal{S}'' \cup \{r_{j}\}}(\alpha)$, then
$$
\phi_{\sigma}^{\alpha}(r_{i})>\phi_{\sigma}^{\alpha}(r_{j}).
$$
\end{proposition}
\begin{proof}
For $\mathcal{S}'$ we have $\sigma_{\mathcal{S}' \cup \{r_{i}, r_{j}\}}(\alpha)-\sigma_{\mathcal{S}' \cup \{r_{j}\}}(\alpha)>\sigma_{\mathcal{S}' \cup \{r_{i}, r_{j}\}}(\alpha)-\sigma_{\mathcal{S}' \cup \{r_{i}\}}(\alpha)$; for any $\mathcal{S}''$ we have $\sigma_{\mathcal{S}'' \cup \{r_{i}, r_{j}\}}(\alpha)-\sigma_{\mathcal{S}'' \cup \{r_{j}\}}(\alpha) \geq \sigma_{\mathcal{S}'' \cup \{r_{i}, r_{j}\}}(\alpha)-\sigma_{\mathcal{S}'' \cup \{r_{i}\}}(\alpha)$. According to Definition~\ref{def_RAE}, we sum up $\mathcal{S}'$ and $\mathcal{S}''$, then we have $\sum_{\mathcal{S} \subseteq \mathcal{R} \setminus \{r_{i}\}} \frac{(\left| \mathcal{R} \right| - \left| \mathcal{S} \right| -1)!\left| \mathcal{S} \right|!}{\left| \mathcal{R} \right|!} \left[ \sigma_{\mathcal{S} \cup \{r_{i}\}}(\alpha)-\sigma_{\mathcal{S}}(\alpha)\right]>\sum_{\mathcal{S} \subseteq \mathcal{R} \setminus \{r_{j}\}} \frac{(\left| \mathcal{R} \right| - \left| \mathcal{S} \right| -1)!\left| \mathcal{S} \right|!}{\left| \mathcal{R} \right|!} \left[ \sigma_{\mathcal{S} \cup \{r_{j}\}}(\alpha)-\sigma_{\mathcal{S}}(\alpha)\right]$, hence $\phi_{\sigma}^{\alpha}(r_{i})>\phi_{\sigma}^{\alpha}(r_{j})$, which completes the proof.
\end{proof}

\subsection{Argumentative Properties} 
\begin{property}[Sign Correctness]
\label{property_sign_correctness}
Let $\alpha \in \mathcal{A}$ and $r \in \mathcal{R}$.
    \begin{enumerate}
        \item If $r \in \mathcal{R^{-}}$, then $\phi_{\sigma}^{\alpha}(r) \leq 0$;
        \item If $r \in \mathcal{R^{+}}$, then $\phi_{\sigma}^{\alpha}(r) \geq 0$.
    \end{enumerate}
\end{property}

\begin{proposition}
\label{proposition_sign_correct_direct}
Let $r$ be a direct edge w.r.t. $\alpha$. 
$\phi_{\sigma}^{\alpha}(r)$ satisfies sign correctness if $\sigma$ satisfies monotonicity.
\end{proposition}
\begin{proof}
For any $\mathcal{S} \subseteq \mathcal{R} \setminus \{r\}$, we have $\frac{(\left| \mathcal{R} \right| - \left| \mathcal{S} \right| -1)!\left| \mathcal{S} \right|!}{\left| \mathcal{R} \right|!}>0$. Since $\sigma$ satisfies monotonicity,  $r \in \mathcal{R}^{-}$ implies  $\sigma_{\mathcal{S} \cup \{r\}}(\alpha) \leq \sigma_{\mathcal{S}}(\alpha)$. According to Definition~\ref{def_RAE}, $\phi_{\sigma}^{\alpha}(r) \leq 0$. Analogously, if $r \in \mathcal{R}^{+}$, then $\phi_{\sigma}^{\alpha}(r) \geq 0$.
\end{proof}

\begin{proposition}
\label{proposition_sign_correct_indirect}
Let $r$ be an indirect edge w.r.t. $\alpha$. Suppose the path sequence from $r$ to $\alpha$ is $
r, r_{1}, \cdots, r_{n} 
(n \geq 1)$. Let $\lambda= \left| \{r_{1}, \cdots, r_{n} \} \cap \mathcal{R}^{-} \right|$. Then the following statements hold if $\sigma$ satisfies monotonicity.
    \begin{enumerate}
        \item If $r \in \mathcal{R^{-}}$ and $\lambda$ is odd, then $\phi_{\sigma}^{\alpha}(r) \geq 0$;
        \item If $r \in \mathcal{R^{-}}$ and $\lambda$ is even, then $\phi_{\sigma}^{\alpha}(r) \leq 0$;
        \item If $r \in \mathcal{R^{+}}$ and $\lambda$ is odd, then $\phi_{\sigma}^{\alpha}(r) \leq 0$;
        \item If $r \in \mathcal{R^{+}}$ and $\lambda$ is even, then $\phi_{\sigma}^{\alpha}(r) \geq 0$.
    \end{enumerate}
\end{proposition}
\begin{proof}
Case 1 (If $r \in \mathcal{R^{-}}$ and $\lambda$ is odd): for any $\mathcal{S} \subseteq \mathcal{R}$, if $\{r_{1}, \cdots, r_{n} \} \nsubseteq \mathcal{S}$, then 
$\sigma_{\mathcal{S} \cup \{r\}}(\alpha) = \sigma_{\mathcal{S}}(\alpha) $ as the path from $r$ to $\alpha$ remains broken, that is $\sigma_{\mathcal{S} \cup \{r\}}(\alpha)-\sigma_{\mathcal{S}}(\alpha)=0$; if $\{r_{1}, \cdots, r_{n} \} \subseteq \mathcal{S}$, then $r$ will not decrease $\sigma(\alpha)$ if $\sigma$ satisfies \emph{monotonicity}, 
because $r$ passes an odd number of attacks on the path to $\alpha$, hence $\sigma_{\mathcal{S} \cup \{r\}}(\alpha)-\sigma_{\mathcal{S}}(\alpha) \geq 0$. Therefore, $\phi_{\sigma}^{\alpha}(r) = \sum_{\mathcal{S} \subseteq \mathcal{R} \setminus \{r\}} \frac{(\left| \mathcal{R} \right| - \left| \mathcal{S} \right| -1)!\left| \mathcal{S} \right|!}{\left| \mathcal{R} \right|!} \left[ \sigma_{\mathcal{S} \cup \{r\}}(\alpha)-\sigma_{\mathcal{S}}(\alpha)\right] \geq 0$. 
Analogously, 
if $r \in \mathcal{R^{-}}$ and $\lambda$ is even, then $\sigma_{\mathcal{S} \cup \{r\}}(\alpha) \leq \sigma_{\mathcal{S}}(\alpha)$ thus $\phi_{\sigma}^{\alpha}(r) \leq 0$;
if $r \in \mathcal{R^{+}}$ and $\lambda$ is odd, then $\sigma_{\mathcal{S} \cup \{r\}}(\alpha) \leq \sigma_{\mathcal{S}}(\alpha)$ thus $\phi_{\sigma}^{\alpha}(r) \leq 0$;
if $r \in \mathcal{R^{+}}$ and $\lambda$ is even, then $\sigma_{\mathcal{S} \cup \{r\}}(\alpha) \geq \sigma_{\mathcal{S}}(\alpha)$ thus $\phi_{\sigma}^{\alpha}(r) \geq 0$.
\end{proof}

\begin{proposition}
\label{proposition_sign_correct_multifold}
Let $r$ be a multifold edge w.r.t. $\alpha$.
$\phi_{\sigma}^{\alpha}(r)$ may violate sign correctness even if $\sigma$ satisfies monotonicity.
\end{proposition}
\begin{proof}
    A counterexample is shown in Figure~\ref{fig_color}, where $(\zeta,\delta) \in \mathcal{R^{-}}$ is a multifold edge w.r.t. $\alpha$, but $\phi_{\sigma^{DF}}^{\alpha}((\zeta,\delta)) \approx 0.0307 > 0$.
\end{proof}

\begin{property}[Counterfactuality]
\label{property_counterfactuality}
Let $\alpha \in \mathcal{A}$ and $r \in \mathcal{R}$.

\begin{enumerate}
    \item If $\phi_{\sigma}^{\alpha}(r) < 0$, then $\sigma(\alpha) \leq
 \sigma_{\mathcal{R} \setminus \{ r \}}(\alpha)$;
    \item If $\phi_{\sigma}^{\alpha}(r) > 0$, then $\sigma(\alpha) \geq \sigma_{\mathcal{R} \setminus \{ r \}}(\alpha)$.
\end{enumerate}
\end{property}

\begin{proposition}
\label{proposition_counterfactuality_direct}
Let $r$ be a direct or indirect edge w.r.t. $\alpha$.
$\phi_{\sigma}^{\alpha}(r)$ satisfies counterfactuality  if $\sigma$ satisfies monotonicity.
\end{proposition}
\begin{proof}
\textbf{Direct:}
According to Proposition~\ref{proposition_sign_correct_direct}, we have $r \in \mathcal{R^{-}} \Rightarrow \phi_{\sigma}^{\alpha}(r) \leq 0$, hence $\phi_{\sigma}^{\alpha}(r) > 0 \Rightarrow r \notin \mathcal{R^{-}}$, that is, $\phi_{\sigma}^{\alpha}(r) > 0 \Rightarrow r \in \mathcal{R^{+}}$. Since we assume monotonicity of $\sigma$, if $r \in \mathcal{R^{+}}$, then $\sigma(\alpha) \geq
 \sigma_{\mathcal{R} \setminus \{ r \}}(\alpha)$.
 Analogously, if $\phi_{\sigma}^{\alpha}(r) < 0$, then $\sigma(\alpha) \leq \sigma_{\mathcal{R} \setminus \{ r \}}(\alpha)$. \\
\textbf{Indirect:}
 According to Proposition~\ref{proposition_sign_correct_indirect}, When $\lambda$ is even, $r \in \mathcal{R^{-}} \Rightarrow \phi_{\sigma}^{\alpha}(r) \leq 0$; $r \in \mathcal{R^{+}} \Rightarrow \phi_{\sigma}^{\alpha}(r) \geq 0$. Therefore, $\phi_{\sigma}^{\alpha}(r) > 0 \Rightarrow r \in \mathcal{R^{+}}$; $\phi_{\sigma}^{\alpha}(r) < 0 \Rightarrow r \in \mathcal{R^{-}}$. Analogously, 
    when $\lambda$ is odd, we have the conclusion $\phi_{\sigma}^{\alpha}(r) > 0 \Rightarrow r \in \mathcal{R^{-}}$; $\phi_{\sigma}^{\alpha}(r) < 0 \Rightarrow r \in \mathcal{R^{+}}$. 
    Therefore, if $\phi_{\sigma}^{\alpha}(r) > 0$, then it means either $r \in \mathcal{R^{+}}$ and $\lambda$ is even, or $r \in \mathcal{R^{-}}$ and $\lambda$ is odd. 
    In both cases, according to the proof in Proposition~\ref{proposition_sign_correct_indirect}, we know $\sigma(\alpha) \geq \sigma_{\mathcal{R} \setminus \{ r \}}(\alpha)$. 
    Similarly, if $\phi_{\sigma}^{\alpha}(r) < 0$, then it means either $r \in \mathcal{R^{-}}$ and $\lambda$ is even, or $r \in \mathcal{R^{+}}$ and $\lambda$ is odd. 
    In both cases, according to the proof in Proposition~\ref{proposition_sign_correct_indirect}, we know $\sigma(\alpha) \leq \sigma_{\mathcal{R} \setminus \{ r \}}(\alpha)$. 
\end{proof}



\begin{proposition}
\label{proposition_counterfactuality_multifold}
Let $r$ be a multifold edge w.r.t. $\alpha$.
$\phi_{\sigma}^{\alpha}(r)$ may violate counterfactuality  even if $\sigma$ satisfies monotonicity.
\end{proposition}
\begin{proof}
    A counterexample is shown in Figure~\ref{fig_property_cf_qua_quan}. 
    $\sigma$ is given by DF-QuAD since it satisfies monotonicity. 
    Let us set $\tau(\mu)=0.4$ and $\tau(\alpha)=\tau(\beta)=\tau(\gamma)=\tau(\zeta)=\tau(\rho)=0$.
    $(\mu,\rho)$ is a multifold edge w.r.t. $\alpha$ because it passes $3$ (more than $1$) paths to reach $\alpha$.
    By DF-QuAD semantics, we have $\sigma^{DF}(\alpha)=0.384$.
    By Definition~\ref{def_RAE}, we have $\phi_{\sigma^{DF}}^{\alpha}((\mu,\rho)) \approx 0.0458 > 0$.
    If $(\mu,\rho)$ is removed, then we apply DF-QuAD again and we have $\sigma^{DF}_{\mathcal{R} \setminus \{ (\mu,\rho) \}}(\alpha)=0.4$.
    Therefore, when $\phi_{\sigma^{DF}}^{\alpha}((\mu,\rho)) > 0$, $\sigma^{DF}(\alpha)=0.384 < \sigma^{DF}_{\mathcal{R} \setminus \{ (\mu,\rho) \}}(\alpha)=0.4$, which violates counterfactuality.
\begin{figure}[ht]
    \centering
    \includegraphics[width=0.25\columnwidth]{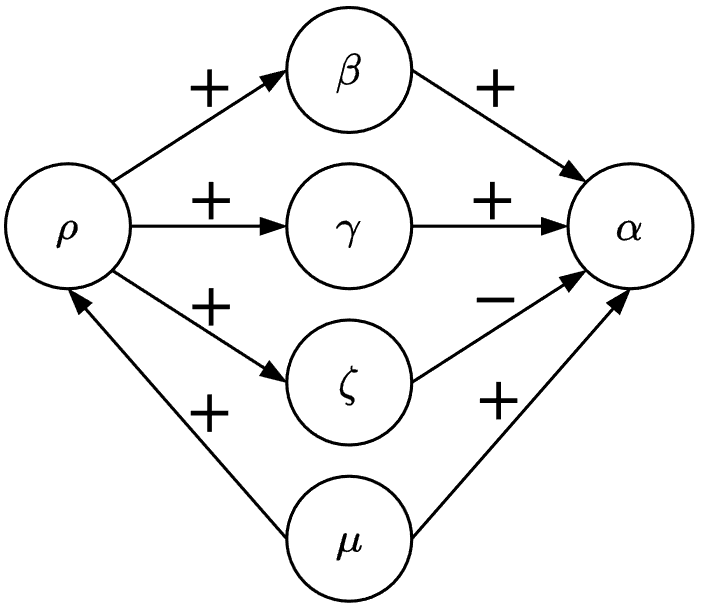}
    \caption{A counterexample.}
    \label{fig_property_cf_qua_quan}
\end{figure}
\end{proof}

\begin{property}[Qualitative Invariability]
\label{property_qua_inv}
Let $\alpha,\beta \in \mathcal{A}$ and $r \in \mathcal{R(\beta)}$.
Let $\phi_{\delta}$ denote $\phi_{\sigma}^{\alpha}(r)$ when setting $\tau(\beta)$ to some $\delta \in \mathbb{I}$.
\begin{enumerate}
    \item If $\phi_{\sigma}^{\alpha}(r) < 0$, then $\forall \delta \in \mathbb{I}$, $\phi_{\delta} \leq 0$;
    \item If $\phi_{\sigma}^{\alpha}(r) > 0$, then $\forall \delta \in \mathbb{I}$, $\phi_{\delta} \geq 0$.
\end{enumerate}
\end{property}

\begin{proposition}
\label{proposition_qua_inv_direct}
Let $r$ be a direct or indirect edge w.r.t. $\alpha$.
$\phi_{\sigma}^{\alpha}(r)$ satisfies qualitative invariability if $\sigma$ satisfies monotonicity.
\end{proposition}
\begin{proof}
\textbf{Direct:}
    According to the proof in Proposition~\ref{proposition_counterfactuality_direct}, we have $\phi_{\sigma}^{\alpha}(r) > 0 \Rightarrow r \in \mathcal{R^{+}}$. 
    According to Proposition~\ref{proposition_sign_correct_direct}, we have $r \in \mathcal{R^{+}} \Rightarrow \phi_{\sigma}^{\alpha}(r) \geq 0$.
    Therefore, if $\phi_{\sigma}^{\alpha}(r) > 0$, then $\forall \delta \in \mathbb{I}$, $\phi_{\delta} \geq 0$. Analogously, if $\phi_{\sigma}^{\alpha}(r) < 0$, then $\forall \delta \in \mathbb{I}$, $\phi_{\delta} \leq 0$.\\
\textbf{Indirect:}
According to the proof in Proposition~\ref{proposition_counterfactuality_direct}, we have: 
    \begin{enumerate}
        \item $\phi_{\sigma}^{\alpha}(r) < 0 \Rightarrow$ $(r \in \mathcal{R^{-}} \wedge \lambda$ is even$)$ $\vee$ $(r \in \mathcal{R^{+}} \wedge \lambda$ is odd$)$;
        \item $\phi_{\sigma}^{\alpha}(r) > 0 \Rightarrow$ $(r \in \mathcal{R^{-}} \wedge \lambda$ is odd$)$ $\vee$ $(r \in \mathcal{R^{+}} \wedge \lambda$ is even$)$.
    \end{enumerate}
    In case 1, according to Proposition~\ref{proposition_sign_correct_indirect}, we have $(r \in \mathcal{R^{-}} \wedge \lambda$ is even$)$ $\Rightarrow \phi_{\sigma}^{\alpha}(r) \leq 0$ and $(r \in \mathcal{R^{+}} \wedge \lambda$ is odd$)$ $\Rightarrow \phi_{\sigma}^{\alpha}(r) \leq 0$. Therefore, in both case, $\forall \delta \in \mathbb{I}$, $\phi_{\delta} \leq 0$.
    Analogously to the proof in case 2.
\end{proof}


\begin{proposition}
\label{proposition_qua_inv_multifold}
Let $r$ be a multifold edge w.r.t. $\alpha$.
$\phi_{\sigma}^{\alpha}(r)$ may violate qualitative invariability even if $\sigma$ satisfies monotonicity.
\end{proposition}
\begin{proof}
    A counterexample is shown in Figure~\ref{fig_property_cf_qua_quan}. 
    $\sigma$ is given by DF-QuAD since it satisfies monotonicity. 
    Let us set $\tau(\mu)=0.4$ and $\tau(\alpha)=\tau(\beta)=\tau(\gamma)=\tau(\zeta)=\tau(\rho)=0$.
    $(\mu,\rho)$ is a multifold edge w.r.t. $\alpha$ because it passes $3$ (more than $1$) paths to reach $\alpha$.
    By Definition~\ref{def_RAE}, we have $\phi_{\sigma^{DF}}^{\alpha}((\mu,\rho)) \approx 0.0458 > 0$.
    Let $\delta=0.7$, then we have $\phi_{\delta} \approx -0.0283 < 0$. 
    Since $\phi_{\sigma^{DF}}^{\alpha}((\mu,\rho))$ does not keep positive for any $\delta$, thus qualitative invariability is violated.
\end{proof}

\begin{property}[Quantitative Variability]
\label{property_variability}
Let $\alpha,\beta \in \mathcal{A}$, $\alpha \neq \beta$, and $\mathcal{R(\beta)}=\{r\}$.
Let $\phi_{\delta}$ denote $\phi_{\sigma}^{\alpha}(r)$ when setting $\tau(\beta)$ to some $\delta \in \mathbb{I}$.
\begin{enumerate}
    \item If $\delta < \tau(\beta)$, then $\left| \phi_{\delta} \right| \leq \left| \phi_{\sigma}^{\alpha}(r) \right|$;
    \item If $\delta > \tau(\beta)$, then $\left| \phi_{\delta} \right| \geq \left| \phi_{\sigma}^{\alpha}(r) \right|$.
\end{enumerate}
\end{property}

\begin{proposition}
\label{proposition_quan_var_direct}
Let $r$ be a direct or indirect edge w.r.t. $\alpha$.
$\phi_{\sigma}^{\alpha}(r)$ satisfies quantitative variability if $\sigma$ satisfies monotonicity.
\end{proposition}
\begin{proof}
\textbf{Direct:}
    We let $\sigma_{\mathcal{S}}^{\delta}(\alpha)$ denote the strength of $\alpha$ in $\mathcal{Q}^{|_\mathcal{S}}$ when setting $\tau(\beta)$ to some $\delta \in \mathbb{I}$.\\
    Suppose $r \in \mathcal{R^{+}}$. 
    As we assume monotonicity of $\sigma$, for any $\mathcal{S} \subseteq \mathcal{R} \setminus \{r\}$,
    if $\delta < \tau(\beta)$, then $\sigma_{\mathcal{S} \cup \left\{ r \right\}}^{\delta}(\alpha) \leq \sigma_{\mathcal{S} \cup \left\{ r \right\}}(\alpha)$.  Since $\sigma_{\mathcal{S}}^{\delta}(\alpha) = \sigma_{\mathcal{S}}(\alpha)$, according to Definition~\ref{def_RAE}, we have $0 < \phi_{\delta} \leq \phi_{\sigma}^{\alpha}(r)$, i.e., $\left| \phi_{\delta} \right| \leq \left| \phi_{\sigma}^{\alpha}(r) \right|$.\\
    Suppose $r \in \mathcal{R^{-}}$. 
    By monotonicity of $\sigma$, for any $\mathcal{S} \subseteq \mathcal{R} \setminus \{r\}$
    if $\delta < \tau(\beta)$, then $\sigma_{\mathcal{S} \cup \left\{ r \right\}}^{\delta}(\alpha) \geq \sigma_{\mathcal{S} \cup \left\{ r \right\}}(\alpha)$. According to Definition~\ref{def_RAE}, since $\sigma_{\mathcal{S}}^{\delta}(\alpha) = \sigma_{\mathcal{S}}(\alpha)$, we have $0 > \phi_{\delta} \geq \phi_{\sigma}^{\alpha}(r)$, i.e., $\left| \phi_{\delta} \right| \leq \left| \phi_{\sigma}^{\alpha}(r) \right|$.
    Therefore, if $\delta < \tau(\beta)$, then $\left| \phi_{\delta} \right| \leq \left| \phi_{\sigma}^{\alpha}(r) \right|$. Analogously, if $\delta > \tau(\beta)$, then $\left| \phi_{\delta} \right| \geq \left| \phi_{\sigma}^{\alpha}(r) \right|$.\\
\textbf{Indirect:}
We let $\sigma_{\mathcal{S}}^{\delta}(\alpha)$ denote the strength of $\alpha$ in $\mathcal{Q}^{|_\mathcal{S}}$ when setting $\tau(\beta)$ to some $\delta \in \mathbb{I}$.
\begin{enumerate}
\item Suppose $r \in \mathcal{R^{+}}$ and $\lambda$ is even. 
Under the guarantee of monotonicity, for any $\mathcal{S} \subseteq \mathcal{R} \setminus \{r\}$, 
if $\delta < \tau(\beta)$, then $\sigma_{\mathcal{S} \cup \left\{ r \right\}}^{\delta}(\alpha) \leq \sigma_{\mathcal{S} \cup \left\{ r \right\}}(\alpha)$. 
Since $\sigma_{\mathcal{S}}^{\delta}(\alpha) = \sigma_{\mathcal{S}}(\alpha)$, according to Definition~\ref{def_RAE}, we have $0 < \phi_{\delta} \leq \phi_{\sigma}^{\alpha}(r)$, i.e., $\left| \phi_{\delta} \right| \leq \left| \phi_{\sigma}^{\alpha}(r) \right|$.
\item Suppose $r \in \mathcal{R^{-}}$ and $\lambda$ is odd. 
Under the guarantee of monotonicity, for any $\mathcal{S} \subseteq \mathcal{R} \setminus \{r\}$, 
if $\delta < \tau(\beta)$, then $\sigma_{\mathcal{S} \cup \left\{ r \right\}}^{\delta}(\alpha) \leq \sigma_{\mathcal{S} \cup \left\{ r \right\}}(\alpha)$. 
Since $\sigma_{\mathcal{S}}^{\delta}(\alpha) = \sigma_{\mathcal{S}}(\alpha)$, according to Definition~\ref{def_RAE}, we have $0 < \phi_{\delta} \leq \phi_{\sigma}^{\alpha}(r)$, i.e., $\left| \phi_{\delta} \right| \leq \left| \phi_{\sigma}^{\alpha}(r) \right|$.
\item Suppose $r \in \mathcal{R^{+}}$ and $\lambda$ is odd.  
Under the guarantee of monotonicity, for any $\mathcal{S} \subseteq \mathcal{R} \setminus \{r\}$
if $\delta < \tau(\beta)$, then $\sigma_{\mathcal{S} \cup \left\{ r \right\}}^{\delta}(\alpha) \geq \sigma_{\mathcal{S} \cup \left\{ r \right\}}(\alpha)$. Since $\sigma_{\mathcal{S}}^{\delta}(\alpha) = \sigma_{\mathcal{S}}(\alpha)$, according to Definition~\ref{def_RAE}, we have  $0 > \phi_{\delta} \geq \phi_{\sigma}^{\alpha}(r)$, i.e., $\left| \phi_{\delta} \right| \leq \left| \phi_{\sigma}^{\alpha}(r) \right|$.
\item Suppose $r \in \mathcal{R^{-}}$ and $\lambda$ is even.  
Under the guarantee of monotonicity, for any $\mathcal{S} \subseteq \mathcal{R} \setminus \{r\}$
if $\delta < \tau(\beta)$, then $\sigma_{\mathcal{S} \cup \left\{ r \right\}}^{\delta}(\alpha) \geq \sigma_{\mathcal{S} \cup \left\{ r \right\}}(\alpha)$. Since $\sigma_{\mathcal{S}}^{\delta}(\alpha) = \sigma_{\mathcal{S}}(\alpha)$, according to Definition~\ref{def_RAE}, we have  $0 > \phi_{\delta} \geq \phi_{\sigma}^{\alpha}(r)$, i.e., $\left| \phi_{\delta} \right| \leq \left| \phi_{\sigma}^{\alpha}(r) \right|$.
\end{enumerate}
Therefore, if $\delta < \tau(\beta)$, then $\left| \phi_{\delta} \right| \leq \left| \phi_{\sigma}^{\alpha}(r) \right|$. Analogously, if $\delta > \tau(\beta)$, then $\left| \phi_{\delta} \right| \geq \left| \phi_{\sigma}^{\alpha}(r) \right|$.
\end{proof}

\begin{proposition}
\label{proposition_quan_var_multifold}
Let $r$ be a multifold edge w.r.t. $\alpha$.
$\phi_{\sigma}^{\alpha}(r)$ may violate quantitative variability even if $\sigma$ satisfies monotonicity.
\end{proposition}
\begin{proof}
    A counterexample is shown in Figure~\ref{fig_property_4}. 
    $\sigma$ is given by DF-QuAD since it satisfies monotonicity. 
    Let us set $\tau(\mu)=0.8$, $\tau(\alpha)=0.5$ and $\tau(\beta)=\tau(\gamma)=\tau(\zeta)=\tau(\rho)=0$.
    $(\mu,\rho)$ is a multifold edge w.r.t. $\alpha$ because it passes $3$ (more than $1$) paths to reach $\alpha$. 
    By Definition~\ref{def_RAE}, we have $\phi_{\sigma^{DF}}^{\alpha}((\mu,\rho)) \approx -0.0693$.
    Let $\delta=1.0$, then we have $\phi_{\delta} \approx -0.0667$.
    Since $\tau(\mu)$ increases from $0.8$ to $1.0$ while $\left|\phi_{\sigma^{DF}}^{\alpha}((\mu,\rho))\right|$ decreases from $0.0693$ to $0.0667$, thus quantitative variability is violated.

\begin{figure}[ht]
    \centering
    \includegraphics[width=0.3\columnwidth]{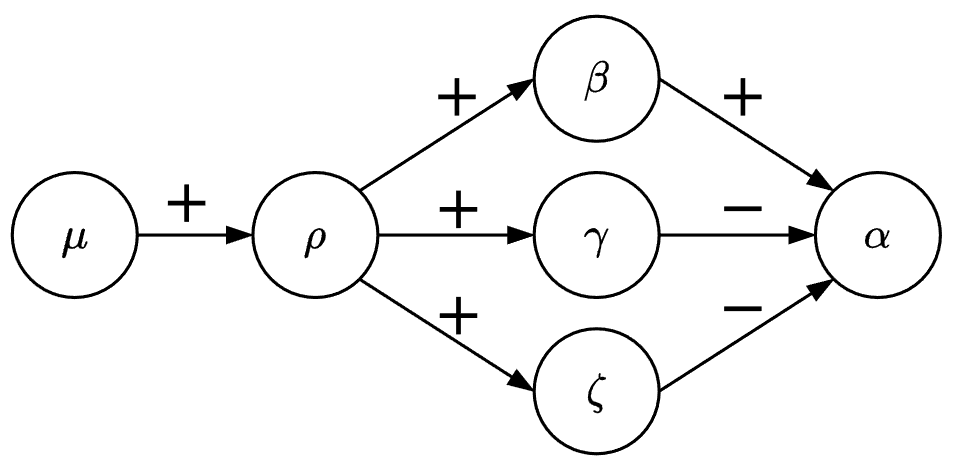}
    \caption{A counterexample.}
    \label{fig_property_4}
\end{figure}    
\end{proof}

\section{Proofs for Section~\ref{sec_approx_algo}} 
\begin{proposition}[Computing RAEs exactly]
\label{proposition_Computational_Complexity_normal}
RAEs can be computed in time $\mathcal{O}(n \cdot 2^{n} \cdot T(m,n))$.
\end{proposition}
\begin{proof}
The strength can be computed in time $T(m,n)$  by assumption.
The exact RAE for one edge can thus be computed in time $\mathcal{O}(2^n\cdot T(m,n))$. Therefore, the total runtime is $\mathcal{O}(n \cdot 2^{n} \cdot T(m,n))$.
\end{proof}


\setcounter{proposition}{16}
\begin{proposition}[Approximating RAEs]
\label{proposition_Computational_Complexity_approx}
If the number of samples for each edge is $N$, then approximate RAEs can be generated in time $\mathcal{O}(n \cdot N \cdot T(m,n))$.
\end{proposition}
\begin{proof}
The strength can be computed in time $T(m,n)$ 
by assumption.
The approximate RAE for each edge can thus be computed in time $\mathcal{O}(N\cdot T(m,n))$. Hence, the total runtime is $\mathcal{O}(n \cdot N \cdot T(m,n))$.
\end{proof}

\section{Additional Details for Section~\ref{sec_intro}}
We show the concrete values of the RAEs in Figure~\ref{fig_introduction}.

In Figure~\ref{fig_introduction}, the base scores are set as
$\tau(\alpha)=0.5$, $\tau(\beta)=0.9$, $\tau(\gamma)=0.1$, $\tau(\delta)=0.1$, $\tau(\xi)=0.1$, $\tau(\eta)=0.1$.\\
Applying DF-QuAD semantics according to Section~\ref{dfquad_details}, we have
$\sigma^{DF}(\alpha)\!=\!0.95095$, $\sigma^{DF}(\beta)\!=\!0.9$, $\sigma^{DF}(\gamma)=0.91$, $\sigma^{DF}(\delta)=0.91$, $\sigma^{DF}(\xi)=0.009$, $\sigma^{DF}(\eta)=0.0099$.\\
We apply RAEs, we have 
$\phi_{\sigma^{DF}}^{\alpha}(r_1) = 0.1981125$, 
$\phi_{\sigma^{DF}}^{\alpha}(r_2) = 0.2471125$, 
$\phi_{\sigma^{DF}}^{\alpha}(r_3) = 0.0135$, 
$\phi_{\sigma^{DF}}^{\alpha}(r_4) = 0.016$, 
$\phi_{\sigma^{DF}}^{\alpha}(r_5) = -0.034$, 
$\phi_{\sigma^{DF}}^{\alpha}(r_6) = -0.011875$, 
$\phi_{\sigma^{DF}}^{\alpha}(r_7) = 0.0221125$.


\section{Additional Details for Section~\ref{sec_preliminaries}}
\begin{example}
\label{example_1_appendix}
Consider the QBAF in Figure~\ref{fig_counter_appendix}. The base scores of all arguments are set to $0.5$, and the strengths of arguments are evaluated by DF-QuAD gradual semantics denoted by $\sigma^{DF}$.

\begin{figure}[h]
    \centering
    \includegraphics[width=0.3\columnwidth]{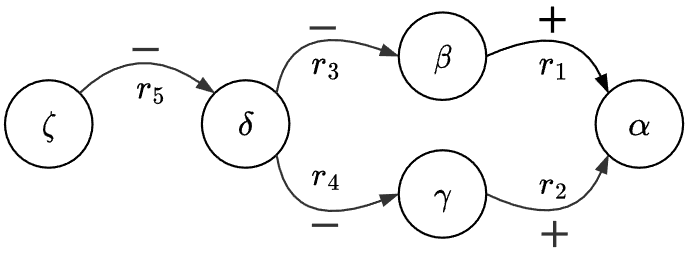}
    \caption{An example of QBAF.}
    \label{fig_counter_appendix}
\end{figure}

According to the definition of DF-QuAD in Section~\ref{dfquad_details}:

For $\zeta$, $v_{\zeta-att}=0$ and $v_{\zeta-sup}=0$, hence $\sigma^{DF}(\zeta)=\tau(\zeta)=0.5$.

For $\delta$, $v_{\delta-att}=0.5$ and $v_{\delta-sup}=0$, hence $\sigma^{DF}(\delta)=\tau(\delta)-\tau(\delta)\cdot(v_{\delta-att}-v_{\delta-sup})=0.25$.

For $\gamma$, $v_{\gamma-att}=0.25$ and $v_{\gamma-sup}=0$, hence $\sigma^{DF}(\gamma)=\tau(\gamma)-\tau(\gamma)\cdot(v_{\gamma-att}-v_{\gamma-sup})=0.375$.

For $\beta$, $v_{\beta-att}=0.25$ and $v_{\beta-sup}=0$, hence $\sigma^{DF}(\beta)=\tau(\beta)-\tau(\beta)\cdot(v_{\beta-att}-v_{\beta-sup})=0.375$.

For $\alpha$, $v_{\alpha-att}=0$ and $v_{\alpha-sup}=1-(1-0.375)(1-0.375)=0.609375$, hence $\sigma^{DF}(\alpha)=\tau(\alpha)+(1-\tau(\alpha))\cdot(v_{\alpha-sup}-v_{\alpha-att})=0.5+0.5\times0.609375=0.8046875$.

To sum up, we have 
$\sigma^{DF}(\alpha)=0.8046875$, 
$\sigma^{DF}(\beta)=0.375$, 
$\sigma^{DF}(\gamma)=0.375$, 
$\sigma^{DF}(\delta)=0.25$, 
$\sigma^{DF}(\zeta)=0.5$.
\end{example}

\section{Additional Experimental Results for Section~\ref{sec_approx_algo}} 
\subsection{Hardware Specifications}
We ran all experiments on a Windows PC (OS: Windows 10 Enterprise, Version: 22H2, 64-bit operating system; Processor: AMD Ryzen 7 PRO 3700 8-Core Processor, 3.59 GHz; Memory: 16.0 GB).

\subsection{Additional Experimental Results}
\begin{figure}[h]
    \centering
    \includegraphics[width=0.7\columnwidth]{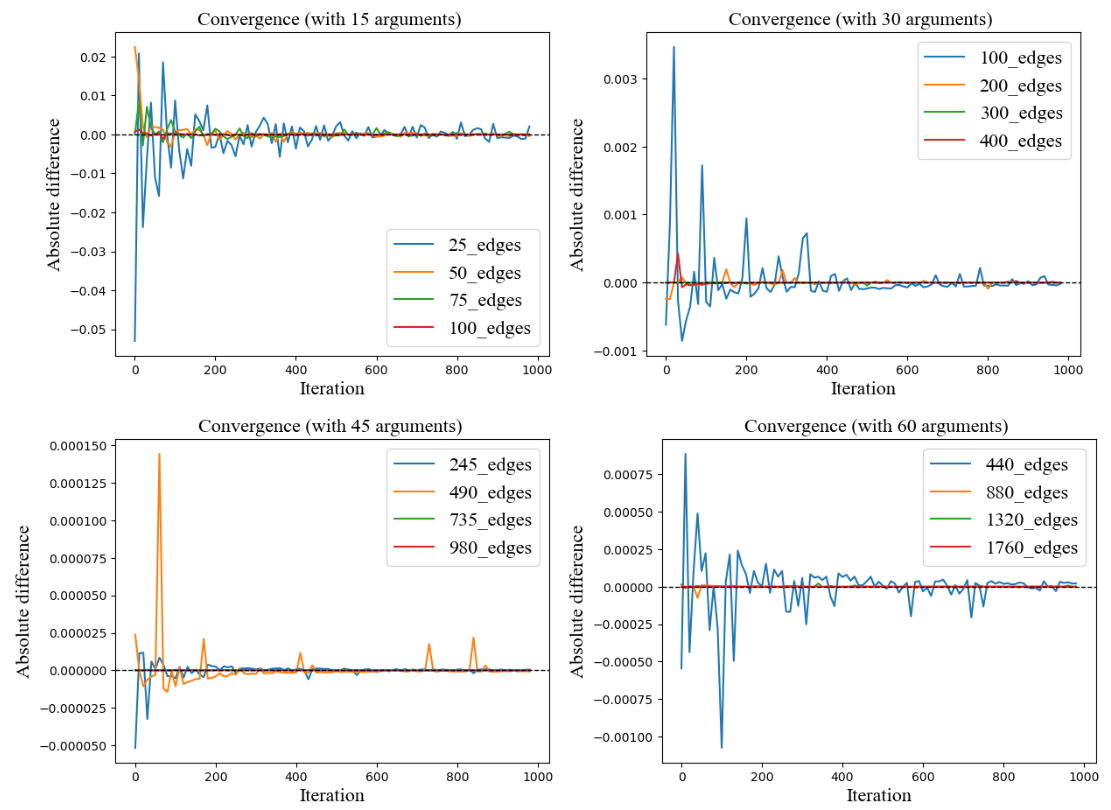}
    \caption{The results are based on random acyclic QBAFs with different number of arguments and edges.}
    \label{fig_convergency_appendix}
\end{figure}

\begin{figure}[h]
    \centering
    \includegraphics[width=0.7\columnwidth]{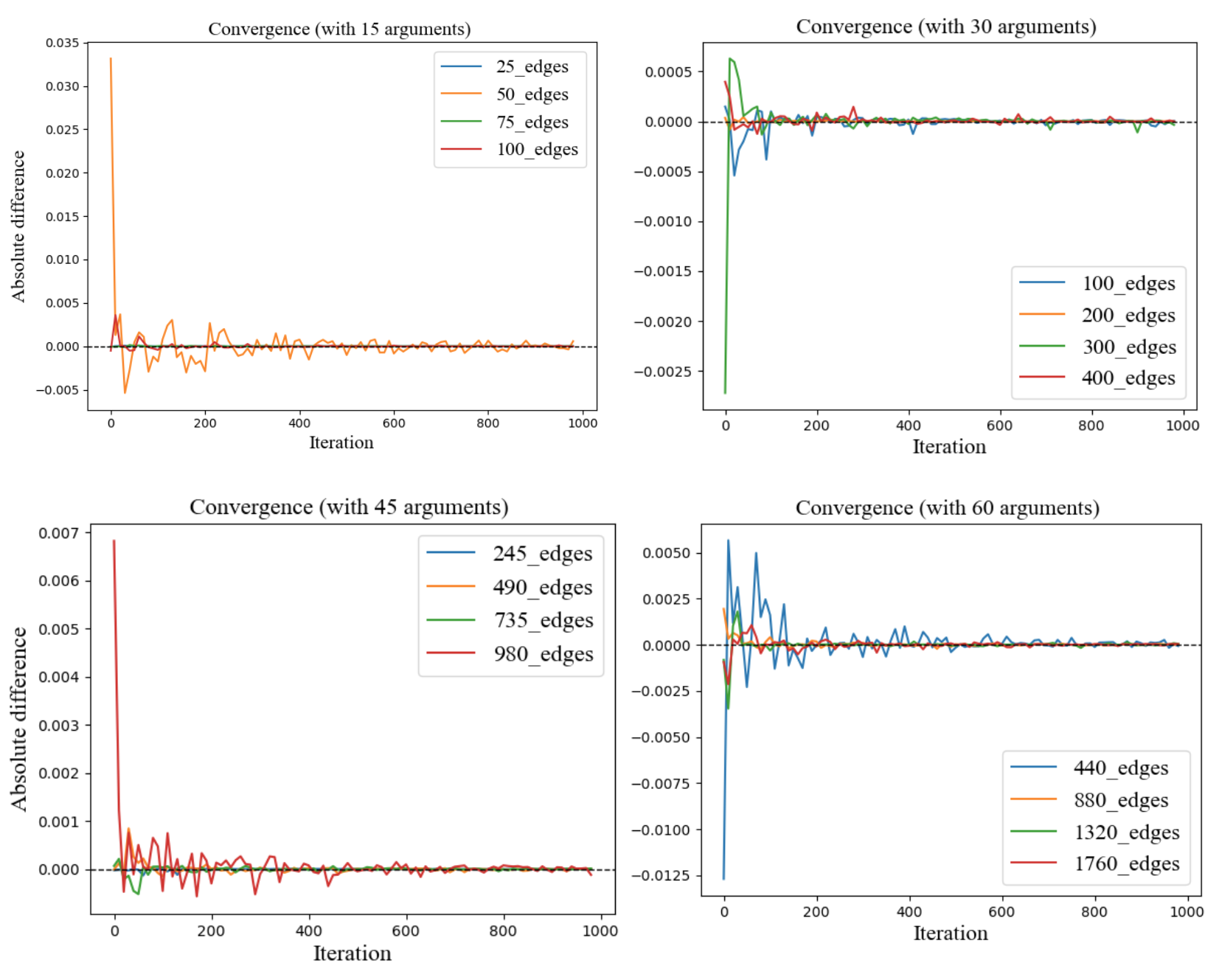}
    \caption{The results are based on random cyclic QBAFs with different number of arguments and edges.}
    \label{fig_convergency_2_appendix}
\end{figure}

Figure~\ref{fig_convergency_appendix} and Figure~\ref{fig_convergency_2_appendix} show additional experimental results of convergence for larger acyclic and cyclic QBAFs, respectively. Figure~\ref{fig_convergency_appendix} is based on DF-QuAD gradual semantics, while Figure~\ref{fig_convergency_2_appendix} is based on QE semantics.\\
Tables~\ref{tab_1}
-
\ref{tab_4} show the runtime in milliseconds of one trial for different sizes of acyclic/cyclic QBAFs.

\begin{table}[!ht]
\caption{Runtime(ms) for random QBAFs with 15 arguments and 25/50/75/100 edges.}
\label{tab_1}
    \centering
    \begin{tabular}{lrrrr}
    \hline
        ~ & 25 & 50 & 75 & 100 \\ \hline
        15 (acyclic) & 0.9 & 1.0 & 1.1 & 1.3 \\
        15 (cyclic) & 14 & 18 & 17 & 23 \\ \hline
    \end{tabular}
\end{table}

\begin{table}[!ht]
\caption{Runtime(ms) for random QBAFs with 30 arguments and 100/200/300/400 edges.}
\label{tab_2}
    \centering
    \begin{tabular}{lrrrr}
    \hline
        ~ & 100 & 200 & 300 & 400 \\ \hline
        30 (acyclic) & 1.5 & 2.2 & 3.2 & 5.6 \\
        30 (cyclic) & 36 & 37 & 47 & 62 \\ \hline
    \end{tabular}
    
\end{table}
\begin{table}[!ht]
\caption{Runtime(ms) for random QBAFs with 45 arguments and 245/490/735/980 edges.}
\label{tab_3}
    \centering
    \begin{tabular}{lrrrr}
    \hline
        ~ & 245 & 490 & 735 & 980 \\ \hline
        45 (acyclic) & 2.5 & 5.7 & 10.4 & 15.3 \\
        45 (cyclic) & 64 & 101 & 127 & 150 \\ \hline
    \end{tabular}
\end{table}

\begin{table}[!ht]
\caption{Runtime(ms) for random QBAFs with 60 arguments and 440/880/1320/1760 edges.}
\label{tab_4}
    \centering
    \begin{tabular}{lrrrr}
    \hline
        ~ & 440 & 880 & 1320 & 1760 \\ \hline
        60 (acyclic) & 5 & 14 & 27 & 43 \\
        60 (cyclic) & 106 & 142 & 216 & 296 \\ \hline
    \end{tabular}
\end{table}

\section{Additional Details for Section~\ref{sec_case_study}} 

\subsection{Case Study 1: Fraud Detection}
We show the contents of the arguments in Figure~\ref{fig_case1_big_AF} in Table~\ref{tab_fraud_arguments}.
\begin{table}[b]
\centering
\begin{tabular}{|c|l|c|l|}
\hline
Arg & Content & Arg & Content \\
\hline
2 & It is a fraud case. & 25 & The basic facts of fabrication stand. \\
3 & It is not a fraud case. & 26 & The victim had abnormal behaviours. \\
4 & The fraud elements are satisfied. & 27 & The victim delivered the property voluntarily. \\
5 & The fraud elements are not satisfied. & 28 & The victim delivered the property consciously. \\
6 & The suspect is with limited capacity. & 29 & The suspect refused to pay refund. \\
7 & The suspect is with incapacity. & 30 & The suspect betrayed the victim's transaction purpose. \\
8 & The suspect is under the age of criminal responsibility. & 31 & The property was transferred to the suspect. \\
9 & The suspect has no capacity of criminal responsibility. & 32 & The collection account is a black account. \\
10 & The suspect had indirect intention. & 33 & The suspects terminated contact with the victim. \\
11 & The suspect fell into negligence. & 34 & The victim's complaint is not credible. \\
12 & The behavior elements of fraud cases are satisfied. & 35 & The transaction is illegal. \\
13 & The result elements of fraud cases are satisfied. & 36 & The transaction occurred between acquaintances. \\
14 & The suspect had direct intention for fraud. & 37 & The suspect faked official materials. \\
15 & The auxiliary facts elements are satisfied. & 38 & The transaction was at variance with business logic. \\
16 & The victim has clear cognition. & 39 & The suspect faked official identity. \\
17 & The suspect fabricates facts. & 40 & The victim belongs to a gullible group. \\
18 & The victim fell into cognitive error. & 41 & The transaction contained high-risk operations. \\
19 & The victim's property has been delivered. & 42 & The transaction was abnormal. \\
20 & The suspect has obtained the victim's property. & 43 & The collection account is a mature account. \\
21 & The suspect had an illegal possession purpose. & 44 & The collection account is an account for daily use. \\
22 & The case is different from common fraud. & 45 & The collection behavior is suspicious. \\
23 & The victim's complaints are malicious. & 46 & The collection account has a high-risk record. \\
24 & The victim had speculative motivations. & 47 & The suspect removed the victim from the communication list. \\
   &                            & 48 & The suspect had no response. \\
\hline
\end{tabular}
\caption{Contents of arguments in Figure~\ref{fig_case1_big_AF}.}
\label{tab_fraud_arguments}
\end{table}
The specific RAEs of edges are shown in Table~\ref{tab_fraud_scores}.
\begin{table*}
\centering
\caption{RAEs in ascending order for the QBAF in Figure~\ref{fig_case1_big_AF_color}.}
\label{tab_fraud_scores}
\begin{tabular}{lclclclc}
\hline
\textbf{Relation} & \textbf{RAE} & \textbf{Relation} & \textbf{RAE} & \textbf{Relation} & \textbf{RAE} & \textbf{Relation} & \textbf{RAE}\\
\hline
att (3,1)  &-4.55E-01 &sup (23,16)  &-1.60E-03 &sup (39,25)	&3.49E-05 &sup (33,21)	&4.95E-04\\
att (5,2)  &-9.13E-02 &sup (34,23)  &-4.74E-04 &sup (41,26)	&3.79E-05 &sup (18,12)	&1.27E-03\\
sup (16,5) &-1.75E-02 &sup (35,24)  &-3.62E-04 &sup (46,32)	&6.24E-05 &sup (19,13)	&1.38E-03\\
sup (8,3)  &-1.64E-02 &sup (36,24)  &-3.52E-04 &sup (48,33)	&6.71E-05 &sup (20,13)	&1.57E-03\\
sup (7,3)  &-1.59E-02 &att (44,32)  &-5.82E-05 &sup (30,20)	&1.40E-04 &sup (17,12)	&1.73E-03\\
sup (11,3) &-1.51E-02 &att (43,32)	&-4.06E-05 &sup (31,20)	&1.47E-04 &sup (21,14)	&1.78E-03\\
sup (6,3)  &-1.51E-02 &sup (37,25)	&1.60E-05  &sup (28,19)	&1.49E-04 &sup (14,4)	&8.41E-03\\
sup (10,3) &-1.46E-02 &sup (42,26)	&2.31E-05  &sup (29,20)	&1.49E-04 &sup (13,4)	&9.36E-03\\
sup (9,3)  &-1.32E-02 &sup (38,25)	&2.56E-05  &sup (27,19)	&2.12E-04 &sup (12,4)	&9.59E-03\\
sup (15,5) &-1.23E-02 &sup (40,26)	&2.78E-05  &sup (32,21)	&2.17E-04 &sup (4,2)	&1.15E-01\\
sup (24,16)&-2.89E-03 &sup (45,32)	&2.79E-05  &sup (25,17)	&2.98E-04 &sup (2,1)	&2.60E-01\\
sup (22,15)&-2.31E-03 &sup (47,33)	&3.09E-05  &sup (26,18)	&3.74E-04\\
\hline
\end{tabular}
\end{table*}

\subsection{Case Study 2: Large Language Models}
According to the definition of the QE semantics in Section~\ref{qe_details}: 

For $\beta$, $E_\beta = 0$, hence $\sigma(\beta)=\tau(\beta)=0.6$.\\
For $\gamma$, $E_\gamma = 0$, hence $\sigma(\gamma)=\tau(\gamma)=0.9$.\\
For $\delta$, $E_\delta = \sigma(\gamma)-\sigma(\beta) = 0.3$, hence $\sigma(\delta)=\tau(\delta)+(1-\tau(\delta))\cdot\frac{E_\delta^2}{1+E_\delta^2}=0.72$.\\
For $\alpha$, $E_\alpha = \sigma(\gamma)+\sigma(\delta)-\sigma(\beta) = 1.02$, hence $\sigma(\alpha)=\tau(\alpha)+(1-\tau(\alpha))\cdot\frac{E_\alpha^2}{1+E_\alpha^2}=0.90$.\\

We next show the exact prompt we used in the case study. However, the response from ChatGPT may vary because the outputs
are generated by a sampling algorithm\footnote{The transfomer in ChatGPT can only predict token probabilities and output sequences have to be generated by repeatedly sampling tokens and
feeding them back into the input. Since the number of tokens is 
in the order of ten thousands, it is impossible to look at all
possible token sequences and token sequences are generated by
randomized heuristic search algorithms.}, etc.\\\\
\textbf{Prompt:}\\
\emph{
Given an argument $\alpha$ : `It is easy for children to learn a foreign language well.'\\
Could you please provide 3 arguments $\beta$, $\gamma$, and $\delta$ that satisfy the following requirements? (each argument should be around 20 words)\\
requirement 1: Provide an argument $\beta$ attacking $\alpha$;\\
requirement 2: Provide an argument $\gamma$ supporting $\alpha$;\\
requirement 3: Provide an argument $\delta$ supporting $\alpha$;\\
requirement 4:  $\beta$  directly attacks $\delta$;\\
requirement 5:  $\gamma$ directly supports $\delta$;\\
requirement 6:  Provide confidence score for $\alpha$, $\beta$, $\gamma$, and $\delta$ as their initial weights ranging from 0 to 1.\\
In the end, give some explanations why the provided arguments satisfy the requirements.\\
}
\textbf{Response:}\\
\emph{
Sure, here are the three arguments as per the requirements:\\
$\alpha$: ``It is easy for children to learn a foreign language well."
Confidence score for $\alpha$: 0.8\\
$\beta$: ``Learning a foreign language requires cognitive maturity, which children lack. Hence, it's difficult for them to excel."
Confidence score for $\beta$: 0.6\\
$\gamma$: ``Studies show that young children possess higher neuroplasticity, making language learning more effective."
Confidence score for $\gamma$: 0.9\\
$\delta$: ``Children immersed in a foreign language environment from an early age have better language acquisition."
Confidence score for $\delta$: 0.7\\
Explanations:\\
1. Argument $\beta$ attacks argument $\alpha$ by emphasizing that children lack the cognitive maturity required to learn a foreign language well. It highlights the challenges faced by children in excelling in language learning, countering the notion that it is easy for them.\\
2. Argument $\gamma$ supports $\alpha$ by citing studies that demonstrate young children's higher neuroplasticity, which refers to the brain's ability to adapt and learn. This evidence suggests that children may be more adept at learning a foreign language than adults.\\
3. Argument $\delta$ also supports $\alpha$ by stating that children exposed to a foreign language environment from an early age have better language acquisition. This implies that children have an advantage in learning a foreign language when immersed in the language consistently.\\
4. Argument $\beta$ directly attacks argument $\delta$ by disregarding the benefits of early exposure to a foreign language and arguing that cognitive maturity is the primary factor affecting language learning.\\
5. Argument $\gamma$ directly supports argument $\delta$ by reinforcing the idea that young children have a higher capacity for language learning, aligning with the notion that early exposure to a foreign language is beneficial.\\
The initial confidence scores assigned to each argument are subjective and reflect the confidence in the validity and strength of the arguments based on available evidence and reasoning.\\
Overall, the provided arguments meet the specified requirements, with one argument attacking $\alpha$ and two arguments supporting it. Additionally, $\beta$ directly challenges $\delta$, while $\gamma$ directly aligns with $\delta$, showcasing a clear connection between the supporting arguments. The confidence scores are arbitrary and represent a subjective assessment of the persuasiveness of each argument.
}
\newpage
\setcounter{page}{1}
\onecolumn
\appendix


\bibliographystyle{named}

\end{document}